\documentclass[10pt,twocolumn,letterpaper]{article}

\usepackage{cvpr}              
\usepackage{amsmath}
\usepackage{amssymb}
\usepackage{rotating}
\usepackage[dvipsnames]{xcolor}
\usepackage{amssymb}
\usepackage{amsmath}
\usepackage{pifont}
\usepackage{multirow}
\usepackage{tabularx}
\usepackage{makecell}
\usepackage{colortbl}
\usepackage{algorithm2e}
\usepackage{algpseudocode}
\usepackage{bm}
\usepackage{gensymb}
\usepackage[dvipsnames]{xcolor}
\usepackage[outline]{contour}
\usepackage{floatrow}
\usepackage[rightcaption]{sidecap}
\usepackage{bbm}
\definecolor{turquoise}{cmyk}{0.65,0,0.1,0.3}
\definecolor{purple}{rgb}{0.65,0,0.65}
\definecolor{dark_green}{rgb}{0, 0.5, 0}
\definecolor{orange}{rgb}{0.8, 0.6, 0.2}
\definecolor{red}{rgb}{0.8, 0.2, 0.2}
\definecolor{darkred}{rgb}{0.6, 0.1, 0.05}
\definecolor{blueish}{rgb}{0.0, 0.3, .6}
\definecolor{light_gray}{rgb}{0.7, 0.7, .7}
\definecolor{pink}{rgb}{1, 0, 1}
\definecolor{greyblue}{rgb}{0.25, 0.25, 1}
\definecolor{LightRed}{rgb}{0.99,0.89,0.89}
\def\eg{\emph{e.g}\onedot} 
\def\ie{\emph{i.e}\onedot} 
\def\cf{\emph{c.f}\onedot}

\def\etal{\emph{et al}\onedot}
\makeatother
\newcommand{\cmark}{\ding{51}}%
\newcommand{\xmark}{\ding{55}}%

\newcommand{\PC}{\mathbf{S}}

\newcommand{\Obj}{\mathbf{X}}
\newcommand{\loss}{\mathcal{L}}
\newcommand{\Fin}{\mathbf{F_{inv}}}
\newcommand{\Feqv}{\mathbf{F_{eqv}}}
\newcommand{\Fcentroid}{\mathbf{F_{c}}}
\newcommand{\Fscale}{\mathbf{F_{s}}}

\newcommand{\Fse}{\mathbf{F}_{se3}}

\DeclareMathOperator*{\argmin}{arg\,min}

\newcommand{\more}{\textsc{MoRE$^2$}}
\definecolor{tabfirst}{rgb}{1, 0.7, 0.7} 
\definecolor{tabsecond}{rgb}{1, 0.85, 0.7} 
\definecolor{tabthird}{rgb}{1, 1, 0.7} 

\definecolor{teaser_gray}{rgb}{0.88, 0.89, 0.90} 
\definecolor{teaser_green}{rgb}{0.77, 0.88, 0.70} 
\definecolor{teaser_blue}{rgb}{0.70, 0.78, 0.90} 
\definecolor{teaser_yellow}{rgb}{1.0, 0.90, 0.60}

\definecolor{cvprblue}{rgb}{0.21,0.49,0.74}
\usepackage[pagebackref=false,breaklinks,colorlinks,citecolor=cvprblue]{hyperref}
\usepackage[capitalize]{cleveref}
\def\paperID{2881} 
\def\confName{CVPR}
\def\confYear{2024}

\title{Living Scenes: Multi-object Relocalization\\ and Reconstruction in Changing 3D Environments}

\author{ Liyuan Zhu\textsuperscript{1} \quad
Shengyu Huang\textsuperscript{2} \quad
Konrad Schindler\textsuperscript{2} \quad
Iro Armeni\textsuperscript{1} 
\quad
\vspace{5px}
\\
{ \textsuperscript{1}{Stanford University} \quad \textsuperscript{2}{ETH Zurich} } \\
}

\begin{document}

\twocolumn[{%
	\renewcommand\twocolumn[1][]{#1}%
	\vspace{-28pt}
        \maketitle
	\begin{center}
	\includegraphics[width=.96\linewidth]{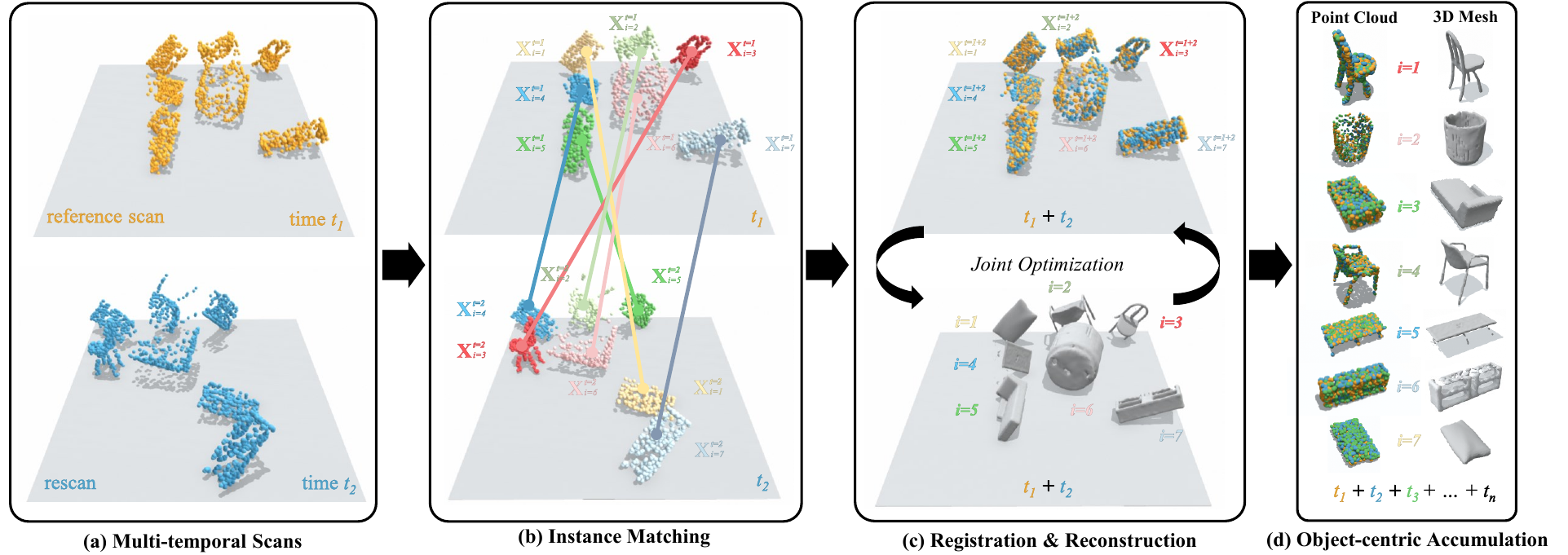}
        \vspace{-1mm}
	\captionof{figure}{\textbf{Living Scenes.} A \textit{living scene} is a 3D environment with multiple moving objects that evolves over time. \textit{(a)} Two temporal observations (scans) represent the scene at times $(t_1, t_2)$ and capture the objects having moved around. To understand the change in the scene, given instance segmentation, we \textit{(b)} match object point clouds from $t_1$ and $t_2$ that belong to the same instance; \textit{(c)} register and reconstruct the matches through our joint optimization, \textit{(d)} accumulate all point clouds per instance from the multiple temporal scans, improving the registration and reconstruction quality over time. We illustrate on two scans for simplicity.}
	\label{fig:teaser}
	\end{center}    
}]

\maketitle
\setlength{\abovedisplayskip}{5pt}
\setlength{\belowdisplayskip}{5pt}
\begin{abstract}
\vspace{-12pt}
Research into dynamic 3D scene understanding has primarily focused on short-term change tracking from dense observations, while little attention has been paid to long-term changes with sparse observations. We address this gap with \textbf{\more}, a novel approach for multi-object relocalization and reconstruction in evolving environments. We view these environments as ``living scenes" and consider the problem of transforming scans taken at different points in time into a 3D reconstruction of the object instances, whose accuracy and completeness increase over time.
At the core of our method lies an SE(3)-equivariant representation in a single encoder-decoder network, trained on synthetic data. This representation enables us to seamlessly tackle instance matching, registration, and reconstruction. We also introduce a joint optimization algorithm that facilitates the accumulation of point clouds originating from the same instance across multiple scans taken at different points in time. We validate our method on synthetic and real-world data and demonstrate state-of-the-art performance in both end-to-end performance and individual subtasks. 
[\href{https://www.zhuliyuan.net/livingscenes}{project}]

\end{abstract}    
\section{Introduction}
\label{sec:intro}
\vspace{-2mm}
3D scene reconstruction serves as the foundation for numerous applications of computer vision and robotics, such as mixed reality, navigation, and embodied perception.
Many of these applications involve the repeated execution of similar tasks, such as cleaning or searching objects, within a given environment. Consequently, they would benefit from an integrated understanding of the environment, accumulated over multiple 3D scans acquired at different points in time.
Such a cumulative scene understanding can enhance interaction with scene objects by progressively improving their geometric completeness and accuracy over time -- especially when previously unseen parts are unveiled -- and could help to develop a foundational understanding of objects' relocation within the environment.
Progressively acquiring a cumulative scene understanding, characterized by increasing geometric completeness and accuracy of constituent objects, can be framed as multi-object relocalization and reconstruction. In this context, object relocalization refers to estimating the 6DoF motion that an object has undergone between two scans, relative to the scene background. 
This resonates with efforts in dynamic scene understanding. The bulk of previous research focused on real-time observation of dynamic scenes~\cite{huang2022dynamic,li2021moltr,wen2023bundlesdf}, where object relocalization boils down to tracking~\cite{ess2008mobile,milan2016mot16,li2021moltr,berclaz2011multiple}. Fewer works address the long-term dynamics of 3D scenes \cite{wald2019rio,adam2022objects,sun2023NSS}, where sensor data cannot be captured constantly but rather at irregular points in time. Due to the long gaps between captures, modeling the objects' intermediate motions is infeasible. Alternative methods~\cite{wald2019rio} solve relocalization via point or object matching, followed by object-wise registration~\cite{sarkar2023sgaligner,huang2021predator}. 
Jointly tackling relocalization and reconstruction~\cite{mescheder2019occupancy,park2019deepsdf,duggal2022mending} over time has been largely overlooked. As we will show in \cref{chapter:experiments}, using separate methods for each task tends to yield increased errors.

To bridge this gap, we introduce \textbf{\more}, a method for \textbf{m}ulti-\textbf{o}bject \textbf{re}localization and \textbf{re}construction of evolving environments over long time spans and from sparse observations, aiming to create a \textit{living scene}. This draws inspiration from the concept of living building information models\footnote{ A building information model is a digital representation of a building in terms of semantically meaningful building parts, including their geometry, attributes, and relations.} \cite{Leite_livingscenes}, which treats buildings as living organisms and aspires to maintain their digital twins throughout their lifespan. As shown in \cref{fig:teaser}(a), our approach takes as input multiple 3D point clouds, acquired at different times and segmented into instances. It addresses the creation of a \textit{living scene} by solving three connected tasks, namely: matching (\cref{fig:teaser}(b)), registering (\cref{fig:teaser}(c) top), and reconstructing (\cref{fig:teaser}(c) bottom) all instances. With a \textbf{single encoder-decoder network} trained on synthetic data, our method is able to solve these tasks for real-world scans. At its heart lies an SE(3)-equivariant representation. Additionally, we introduce an optimization scheme that facilitates the accumulation of point clouds originating from the same instance but different scans (\cref{fig:teaser}(c)). We evaluate \more{} on two long-term living scene datasets, a synthetic one that we generate using 3D object models from ShapeNet~\cite{chang2015shapenet} and the real-world 3RScan~\cite{wald2019rio}, and achieve state-of-the-art performance for both the end-to-end system and each of its subtasks. Our contributions are:
\begin{enumerate}
    \item A new object-centric formulation of parsing an evolving 3D indoor environment as a \textit{living scene}. It involves instance matching, relocalization, and reconstruction.
    \item A novel compact object representation that simultaneously tackles all three tasks. It can generalize to real scenes while being trained on synthetic data only. 
    \item A joint optimization algorithm that progressively improves the performance of the point cloud registration and reconstruction as more data are accumulated. 
\end{enumerate}

\section{Related Work}
\label{related_work}
\paragraph{Dynamic point cloud understanding.} Dynamic scenes mainly consist of multiple moving instances and a static background. Modeling such complex environments usually starts with estimating the per-point motion -- \ie, scene flows~\cite{li2021neural_sceneflow,wu2020pointpwc,vogel20113d,liu2019flownet3d} -- in the scene. Nevertheless, conventional scene flow estimation methods are instance-agnostic, lacking high-level scene representations for downstream tasks related to moving agents. Other methods \cite{huang2021multibodysync,baur2021slim,huang2022dynamic,deng2023banana,song2022ogc} further disentangle the segmentation and motion estimation for multi-body scenes and object articulation. By combining detection with motion models, 3D multi-object tracking methods ~\cite{wu2021track,weng20203dtracking,pang2022simpletrack,zhou2022pttr_tracking} directly obtain object instance motions. This line of work relies on regular and constant observations at a high frequency in a short time horizon, such as those in self-driving car datasets ~\cite{Geiger2013IJRR,caesar2020nuscenes,sun2020waymo}. To study long-term changing indoor environments, researchers captured  3D datasets (3RScan~\cite{wald2019rio}, ReScan~\cite{halber2019rescan}, and NSS~\cite{sun2023NSS}), where changes between observations are more drastic, making tracking-based methods~\cite{li2021moltr,shan2021ellipsdf,zhou2022pttr_tracking} no longer applicable. To understand long-term changes, Adam \etal~\cite{adam2022objects} develop a 3D change detection method using geometric transformation consistency, however they do not have a notion of instances and/or semantics. Halber \etal~\cite{halber2019rescan} build a spatiotemporal model for temporal point clouds to improve instance segmentation, however they do not tackle the task of relocalization or reconstruction. Wald \etal~\cite{wald2019rio} focus on the task of object instance relocalization in two temporal point clouds. They introduce a triplet network for local feature matching, aiming to identify 3D keypoint correspondences between instance patches from two observations. In contrast, our approach goes beyond local matching and infers 6DoF transformations at the entire instance level. It additionally performs reconstruction, which no existing method designed for indoor long-term changing environments addresses.

\vspace{-10pt}
\paragraph{$\mathbf{SO(3)}$-equivariant networks.} Randomly oriented point clouds make coordinate-based networks suffer from inconsistent predictions and poor performance. Such negative effects can be partially alleviated by heavy data augmentation during training~\cite{qi2017pointnet}. Thus, preserving rotations, \ie, $\mathrm{SO(3)}$-equivariance, in point cloud processing networks becomes a desired property. Thomas \etal~\cite{thomas2018tensor} apply spherical harmonics to constrain the network to achieve $\mathrm{SE(3)}$ equivariance. $\mathrm{SO(3)}$-transformers~\cite{fuchs2020se} introduce equivariance to the self-attention mechanism~\cite{vaswani2017attention} and significantly improve the efficiency of~\cite{thomas2018tensor}.
Deng \etal develop Vector Neurons (VNN)~\cite{deng2021vector}, a general framework to make MLP-based networks $\mathrm{SO(3)}$-equivariant by vectorizing scalar neurons. GraphOnet~\cite{chen20223eqgraph} extends vector neurons to $\mathrm{SE(3)}$ equivariance. Assaad \etal~\cite{assaad2022vn_transformer} apply rotation equivariant attention for vector neurons. EFEM~\cite{Lei2023EFEM} uses VNN to store shape priors and performs unsupervised object segmentation through expectation maximization. Deng \etal~\cite{deng2023banana} develop a Banach-fixed point network with inter-part equivariance for object articulation and multi-object segmentation. Yang \etal~\cite{yang2023equivact} extend VNN to policy learning and non-rigid object manipulation in robotics. We utilize VNNs to comprehend the dynamics of multiple objects.

\vspace{-10pt}
\paragraph{Point cloud registration.} Aligning posed point clouds is essential for 3D perception and mapping. Several hand-crafted 3D feature descriptors have been developed for local feature matching, such as  FPFH~\cite{rusu2009fpfh} and SHOT~\cite{tombari2010unique}. Following advancements in deep learning, 3DMatch~\cite{zeng20173dmatch}, PerfectMatch~\cite{gojcic2019perfect}, and RPMNet~\cite{yew2020rpm} focus on learning-based descriptors. Predator~\cite{huang2021predator} introduces attention mechanisms~\cite{vaswani2017attention} into finding 3D correspondences, particularly in low-overlapping regions. In \cite{Yew_2022_REGTR,qin2022geometric}, the use of transformer architectures is refined for superpoint matching. \cite{jin2023qreg,Yu_2023_PEAL,sarkar2023sgaligner} leverage additional prior information like surface curvature, 2D image overlap, and scene structure. 
Apart from the correspondence matching methods above, another line of work is to learn equivariant representations to solve the relative pose.
Wang \etal~\cite{wang2022yoho} develop rotation equivariant descriptors using group equivariant learning~\cite{taco_group_eq}. Yu \etal \cite{Yu_2023_RolTR} develop a rotation invariant transformer method to cope with pose variations in point cloud matching.  Zhu \etal~\cite{zhu2022correspondence} directly solve the pairwise rotation using rotation equivariant~\cite{deng2021vector} embeddings.
Our method also uses equivariant representation but additionally leverages neural implicit surfaces to align two point clouds using test-time optimization.

\vspace{-10pt}
\paragraph{Multi-object reconstruction.}
The recent emergence of neural implicit reconstruction~\cite{park2019deepsdf,chen2019imnet,mescheder2019occupancy} has boosted the performance and flexibility of object reconstruction methods. This is attributed to the learned shape prior and differentiability for test-time optimization~\cite{duggal2022mending}.
FroDo~\cite{runz2020frodo} reconstructs object shapes using detection and multi-view optimization. ELLIPSDF~\cite{shan2021ellipsdf} and ODAM~\cite{li2021odam} further introduce geometric representations, specifically superquadrics, to represent shape primitives and constrain multi-view optimization. Irshad \etal~\cite{irshad2022centersnap,irshad2022shapo,lunayach2023fsd} do not rely on existing detectors but instead develop a single-shot pipeline to regress object pose, shape, and appearance. BundleSDF~\cite{wen2023bundlesdf} focuses on single dynamic objects and generalizes to unknown objects using graph optimization. Our reconstruction is built on top of \cite{park2019deepsdf}. We further introduce joint optimization on shape and pose to aggregate multiple observations for more accurate and complete reconstruction.

\section{Living Scenes}
\label{method}
We define a \textbf{living scene} as a built environment with dynamic and static objects. Its reconstruction occurs cumulatively over time from temporal scans and showcases how it has been \textit{lived}. Our method, $\more$, creates living scenes and is designed to understand the rigid motion and the geometry of objects. It reconstructs each individual 3D object separately, with increased accuracy and completeness as more temporal scans become available (\cf \cref{fig:teaser}(d)). These reconstructions can be seamlessly positioned within the scans acquired over time and used for other tasks, \eg, learning from historical data or creating 3D assets.

\paragraph{Problem Setting.} Consider a collection of scans $\{\PC^t  \}_{t=1}^T$ of a 3D environment captured at irregular intervals. Scan $\PC^t$ represents the environment observed at time $t$ and contains a list of point clouds $\{\Obj_i^t \}_{i=1}^N$. Hereafter, we term object-level point clouds as \textit{point cloud} and scene-level point clouds as \textit{scan}.  We denote the first scan $\PC^1$ as the reference scan and the following ones $\{\PC^t | t>1\}$ as temporal rescans. Our goal is formulated as:
\begin{enumerate}
    \item{\textbf{Multi-object relocalization}}: We aim to compute the 6DoF rigid transformation $\{\mathbf{T}_{i}^{t} \in \mathrm{SE(3)} |~t>1\}$ between the point clouds belonging to the same instance in the reference scan and rescan respectively. Specifically, we formulate relocalization in two steps: matching of point clouds followed by their registration.
    \item{\textbf{Object reconstruction}}: The goal is to reconstruct each instance from the accumulated point clouds $\{\Obj^1\circ\mathbf{T}_{i}^{2}\Obj^2\circ...\circ\mathbf{T}_{i}^{t}\Obj^t~|~t>1\}$, where $\circ$ denotes concatenation operation.
\end{enumerate}
Since our method reasons at the instance level, we assume the availability of instance segmentation masks.\footnote{In \cref{chapter:experiments}, we offer an experiment using predicted instance masks as input to $\more$. Note that $\more$ is agnostic to the semantic labels.}.

\begin{figure*}[t]
    \centering
    \includegraphics[width=\linewidth]{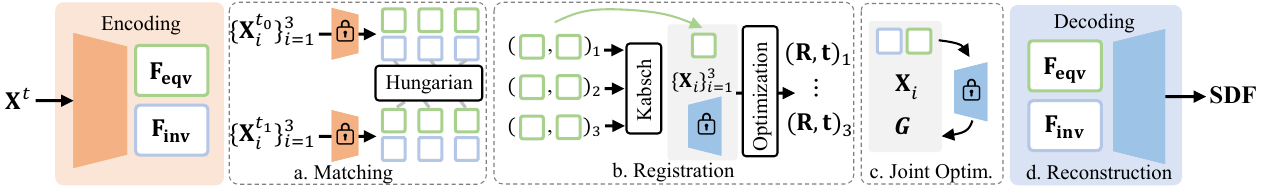}
\vspace{-5px}
    \caption{\textbf{Overview of the $\more$ pipeline.} Given two temporal point clouds with instance masks $\{\Obj^{t_0}_i\}_{i=1}^3$ and $\{\Obj^{t_1}_i\}_{i=1}^3$, we first use the VN encoder to compute the embeddings for each instance. \textit{a)} Matching solves the pairwise correspondences of the same instances using Hungarian matching~\cite{munkres1957hugarian} on the embeddings.  \textit{b)} Registration estimates 6DoF transformations within matched pairs: Kabsch algorithm~\cite{kabsch1976solution} is employed to compute the initial transform, followed by optimization to further refine the registration. \textit{c)} Joint optimization simultaneously refines the registration and \textit{d)} reconstruction. The output is the signed distance values (SDF) of query coordinates.
    }
    \label{fig:overview}
\end{figure*}

\paragraph{Method Overview.}
$\more$ sequentially addresses instance matching (\cref{sec:method_match}), registration (\cref{sec:method_reg}), and reconstruction (\cref{sec:method_recon}) with a single compact representation obtained from our encoder-decoder network~(\cref{method:backbone}), which is trained solely on the object reconstruction task.
We obtain the final accumulated point clouds per instance through joint shape-pose optimization (\cref{sec:joint_optim}). An overview is provided in \cref{fig:overview}.

\subsection{Encoder-decoder Network}
\label{method:backbone}
\paragraph{Vector Neuron Encoder (VN).} 
To obtain rotation equivariant features from point clouds, we follow Deng \etal~\cite{deng2021vector} and ``lift" the neuron representation from a scalar to a vector and preserve the rotation during forward propagation. $\mathrm{SO(3)}$ equivariance and invariance of function $f$ are expressed as:
\begin{equation}
    f(\mathbf{R}\Obj) = \mathbf{R}f(\Obj), \\ f(\mathbf{R}\Obj) = f(\Obj),
\end{equation}
where $\mathbf{R}$ denotes the rotation applied to input point cloud $\mathbf{X}$. \cite{Lei2023EFEM} extends VNNs from $\mathrm{SO(3)}$- to $\mathrm{SIM(3)}$-equivariance by additionally estimating a scale factor and the centroid of the point cloud. The encoder $\Phi$ takes a point cloud $\Obj$ as input and outputs $\mathbf{F} = (\Fin \in \mathbb{R}^{256}, \Feqv \in \mathbb{R}^{3\times 256}, \Fscale  \in \mathbb{R}_{+}, \Fcentroid  \in \mathbb{R}^{3})$, with the four components representing the invariant embedding, equivariant embedding, scale factor, and centroid,  respectively. The $\mathrm{SIM(3)}$-equivariance is then achieved by: 
 \begin{equation}
     \Phi (s\mathbf{R}\Obj+\mathbf{t}) = (\Fin , \mathbf{R}\Feqv, s\Fscale, \Fcentroid + \mathbf{t}).
 \end{equation}
Here $\Phi$ is the VN-encoder and $(s,\mathbf{R},\mathbf{t})\in \mathrm{SIM(3)}$ denote scale, rotation, and translation respectively. The embedding $\mathbf{F}$ is used to canonicalize the query point $\mathbf{p}\in \mathbb{R}^3$ 
\begin{equation}
    \mathbf{F_q} =  \langle \Feqv, (\mathbf{p}-\Fcentroid)/\Fscale \rangle
\end{equation}
where  $\langle \cdot,\cdot \rangle$ denotes channel-wise inner product. The canonicalized feature $\mathbf{F_q}$ is then fed to the decoder.

\vspace{-8pt}
\paragraph{Neural Implicit Decoder.} Here we use DeepSDF~\cite{park2019deepsdf} as our neural implicit decoder. It is an auto-decoder network that takes latent code and query coordinates as input and outputs the SDF value at query location. 
\begin{equation}
\label{eq:sdf}
    \mathbf{SDF}(\mathbf{p}) = \Psi(\Fin, \mathbf{F_q}),
\end{equation}
where $ \Psi$ represents the DeepSDF decoder. 
The latent space trained using DeepSDF decoder lays a solid ground for shape interpolation and test-time optimization~\cite{duggal2022mending}.

\vspace{-8pt}
\paragraph{Training.} We train $\more$ using the L1 reconstruction loss~\cite{park2019deepsdf,Lei2023EFEM} on individual shapes:
\begin{equation}
    \loss_\mathrm{recon} =\frac{1}{K} \sum_{i=1}^{K}|\overline{\mathbf{SDF}}(\mathbf{p}_i) - \mathbf{SDF}(\mathbf{p}_i)|,
\end{equation}
where $\overline{\mathbf{SDF}}(\mathbf{p})$ denotes the ground truth SDF value of $\mathbf{p}$ and $\mathbf{SDF}(\mathbf{p})$ the predicted SDF value. $\mathbf{p}_i$ is the sampled point and $K$ denotes the number of SDF samples.  Additional details are provided in Supp.

Unlike \cite{park2019deepsdf,duggal2022mending}, we make $\more$ category-agnostic by directly training it across multiple classes. Once the training is complete, we freeze the network's weights. Next, we elaborate on how we flexibly adapt the network and embeddings to address the relocalization and reconstruction tasks. 

\subsection{Instance Matching}
\label{sec:method_match}
Given two sets of randomly oriented point clouds 
$\{ (\Obj_i^{t_1})\}_{i=1}^{N},~\{ (\Obj_j^{t_2})\}_{j=1}^{M}$ of size $N$ and $M$, the task is to associate them across time~(\cf \cref{fig:teaser}(b) and \cref{fig:overview}(a)).

We first compute the cosine similarity matrix $\mathbf{\Lambda} \in \mathbb{R}^{N \times M}_{+}$ using all the invariant embedding pairs $\{ \langle\Fin_{i}^{t_1}, \Fin_{j}^{t_2}\rangle \}_{i,j=1}^{N,M}$ as our initial score matrix.
Next, since the equivariant embeddings $\Feqv$ can be treated as 3D coordinates in the latent space given the one-to-one correspondences along the feature dimension, we extract the rotations $\mathbf{R}_{i,j}$ between all equivariant embedding pairs via the Kabsch algorithm~\cite{kabsch1976solution} and consequently factor them out for each pair. Following the factorization, we can compute the alignment residual matrix $\mathbf{E}\in \mathbb{R}^{N\times M}_{+}$, where 
\begin{equation}
    \mathbf{E}(i,j) = ||\mathbf{R}_{i,j} \Feqv_{,i}^{t_1} - \Feqv_{,j}^{t_2} ||_2
\end{equation}
indicates the inverse fitness of equivariant pairs after coarse alignment in the $\mathrm{SO(3)}$ feature space using $\mathbf{R}_{i,j}$. Finally, we compute the aggregated matching score matrix $\mathbf{H} = \mathbf{\Lambda} \oslash \mathbf{E}$, where $\oslash$ denotes element-wise division of matrices. 
Now the problem is to find the assignment that maximizes the total matching score $\sum_{i,j}\mathbf{H}_{i,j}\mathbf{P}_{i,j}$. In light of existing 2D/3D feature matching paradigms~\cite{sarlin2020superglue,gojcic2021weakly}, we use the Hungarian Matching~\cite{munkres1957hugarian} to solve this linear assignment. Considering that the numbers of object instances in two sets can differ and some might remain unmatched after Hungarian Matching, we treat unmatched instances as \textit{removed} or \textit{added} based on their appearance order in time.

\subsection{Instance Registration}
\label{sec:method_reg}
Consider a matched pair $(\Obj^{t_1} \in \mathbb{R}^{3 \times N_1}$, $\Obj^{t_2} \in \mathbb{R}^{3 \times N_2})$  (\cref{fig:teaser}(c) top). The task is to estimate the relative transformation $\mathbf{T}=(\mathbf{R}, \mathbf{t})$ that aligns the source $\Obj^{t_2} $ to the target $\Obj^{t_1}$. To address this, we propose the following optimization-based registration (\cref{fig:overview}(b)). We first compute the $\mathrm{SE(3)}$-equivariant embeddings $\Fse = \Feqv + \Fcentroid$ for each point cloud and solve $(\mathbf{R}, \mathbf{t})$ using Kabsch algorithm~\cite{kabsch1976solution}. This serves as the initialization for our registration. Next, the optimal transformation $(\mathbf{R}^{*}, \mathbf{t}^{*})$ is obtained by minimizing $\loss_{\mathrm{reg}}$:
\begin{equation}
    (\mathbf{R}^{*}, \mathbf{t}^{*}) = \argmin_{(\mathbf{R},\mathbf{t})}\loss_{\mathrm{reg}}(\Obj^{t_1}, \Obj^{t_2}), 
\end{equation}
where $\loss_{\mathrm{reg}}$ is defined as:
\begin{equation}
\label{eq:ref}
\resizebox{0.9\hsize}{!}{$
\begin{aligned}
    \loss_{\mathrm{reg}}(\Obj^{t_1}, \Obj^{t_2}) = \underbrace{||\Psi(\mathbf{F}^{t_1}_\mathbf{inv}, \langle \mathbf{F}^{t_1}_\mathbf{eqv}, (\mathbf{R}_\mathrm{i}\Obj^{t_2}+\mathbf{t}_\mathrm{i}-\mathbf{F}^{t_1}_\mathbf{c})/\mathbf{F}^{t_1}_\mathbf{s} \rangle)||_1}_{\loss_\mathrm{sdf}} \\ + \underbrace{\Tilde{CD}(\mathbf{R}_\mathrm{i}\Obj^{t_2}+\mathbf{t}_\mathrm{i}, \Obj^{t_1})}_{\loss_\mathrm{cd}}.
\end{aligned}$

}
\end{equation}

Here $\loss_\mathrm{sdf}$ denotes the misalignment between the point cloud and zero level-set, and $\loss_\mathrm{cd}$ the chamfer loss between the current estimate and target point cloud.

In the $\mathrm{i}^{th}$ iteration, we use the target embedding $\Fse^{t_1}$ to canonicalize the source point cloud ${\Obj}^{t_2}$ transformed by current $(\mathbf{R}_\mathrm{i}, \mathbf{t}_\mathrm{i})$ and compute $\loss_{\mathrm{reg}}$. We directly optimize $(\mathbf{R}, \mathbf{t})$ through back-propagation on $\mathrm{SE(3)}$ manifold using \emph{TorchLie}~\cite{pineda2022theseus} for faster and more stable convergence~\cite{pan2023panoptic,teed2021tangent}. Our optimization iteratively aligns the source point cloud to the \textit{zero level-set} of the target point cloud, together with minimization of point-wise misalignment.
After optimization, we refine the point cloud alignment using iterative closest point~\cite{besl1992icp} to obtain the final output. Furthermore, we can \textbf{classify static~/dynamic} objects in the scene by thresholding their transformation distances. 

\subsection{Instance Reconstruction}
\label{sec:method_recon}
After obtaining the \textbf{matched} and \textbf{aligned} point cloud pairs $\{(\Obj^{t_1}, \Obj^{t_2}, \mathbf{R}, \mathbf{t})_i\}_{i=0}^M$, we proceed to reconstructing them~(\cref{fig:teaser}(c) bottom). We first down-sample the accumulated point clouds using farthest point sampling (FPS)~\cite{qi2017pointnet++}. Next, we compute the new embedding $\mathbf{F}_{*}$ from the downsampled point cloud. Finally, we query the SDF values of a voxel grid with $64^3$ resolution using $\mathbf{F}_{*}$ and DeepSDF decoder, as is shown in \cref{eq:sdf}. Following the previous literature~\cite{park2019deepsdf,Lei2023EFEM,runz2020frodo}, we use Multi-resolution IsoSurface Extraction (MISE)~\cite{mescheder2019occupancy} to extract the zero level-set as object reconstruction.

\subsection{Joint Optimization for Accumulation}
\label{sec:joint_optim}
So far, we have discussed relocalization and reconstruction between two temporal scans. To leverage observations from multiple scans, we propose a joint optimization algorithm to refine the registration and reconstruction~(\cref{fig:overview}(c),(d)) and accumulate point clouds with increasing geometric accuracy and completeness over time(\cref{fig:teaser}(d)). 

\paragraph{Initialization.} 
Consider the matched and registered point clouds $\{\Obj^{t}\}_{t=t_1}^{t_K}$ and their associated equivariant and invariant embeddings from VN-encoder. For each point cloud $\Obj^{t}$, we compute its $\loss_\mathrm{sdf}$ value and choose the one $\Obj^{*}$ with the best agreement between the point cloud and the zero level set defined by its embeddings, as our initialization. Specifically, we initialize $\mathbf{F}$ with the equivariant embedding $\mathbf{F}_{\textbf{eqv}}^{*}$ and construct the pose graph $\mathbf{G} = \{\mathbf{T}^t\}_{t=t_1}^{t_K}$. Here $\mathbf{T}^t$ aligns $\Obj^t$ to $\Obj^{*}$ and is computed by the previously introduced registration method.

\paragraph{Optimization Objectives.}
We jointly optimize the shared equivariant embedding $\mathbf{F}$ and pose graph $\mathbf{G}$ by minimizing $\loss_{joint}=\loss_\mathrm{sdf}+\loss_z$. Here $ \loss_\mathrm{sdf}$ denotes the SDF error of accumulated point clouds and $\loss_z = ||\mathbf{F}' - \mathbf{F} ||_2$ is the regularization term to constrain variations w.r.t. the initial $\mathbf{F}$. Similar to registration, the pose graph is optimized on $\mathrm{SE(3)}$ manifold~\cite{pineda2022theseus,teed2021tangent} and the embedding is optimized using Adam optimizer~\cite{kingma2014adam} for 200 iterations.

\vspace{0pt}
\subsection{Implementation Details} 
We use the VN Transformer~\cite{assaad2022vn_transformer} and DeepSDF~\cite{park2019deepsdf} as our encoder and decoder, respectively. We implement $\more$ using PyTorch~\cite{paszke2017pytorch} and train it on a single NVIDIA A100 (80GB) GPU for $2\times 10^{5}$ iterations with batch size = 64. We decay the learning rate (0.0001) by 0.3 at $1.2\times 10^{5}$, $1.5\times 10^{5}$, and $1.8\times 10^{5}$ iterations. For more details, we refer the reader to the Supp.

\section{Experiments}
\begin{figure*}[t]
    \centering
    \includegraphics[width=.95\linewidth]{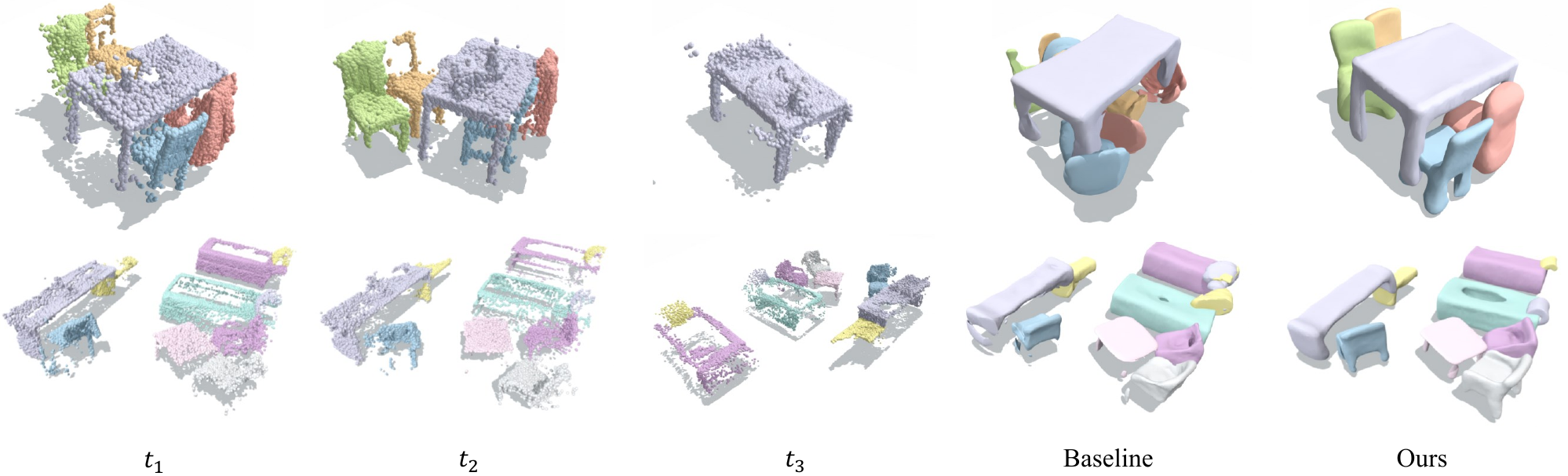}
    \vspace{-5pt}
    \caption{\textbf{End-to-end cumulative reconstruction with multiple scans.} $t_1, t_2$, and $t_3$ denote the same scene captured at three times. Point clouds from $t_2$ and $t_3$ are accumulated to $t_1$. Interestingly, chairs in $t_3$ (top) are removed from the scene, but $\more$ is able to handle it.}
\label{fig:end_to_end}
\end{figure*}
\begin{figure*}[t]
    \centering
    \includegraphics[width=\textwidth]{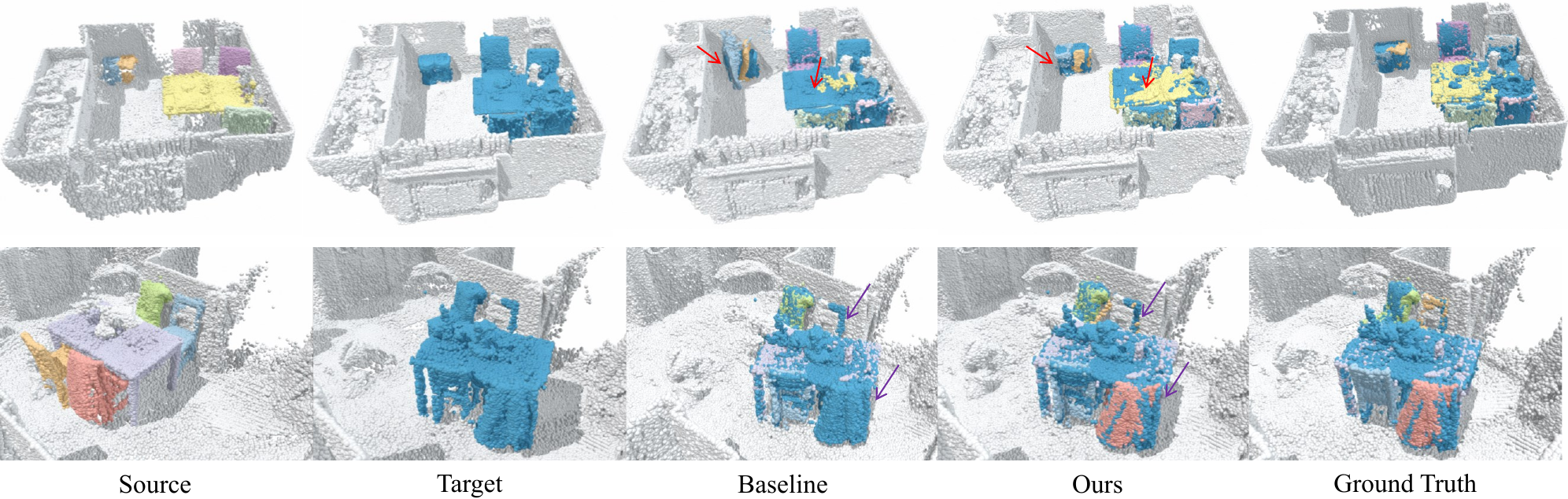}
\vspace{-5mm}
    \caption{\textbf{Multi-object relocalization on \emph{3RScan}~\cite{wald2019rio}.} Instances, uniquely colored in source scan, are matched and registered to their corresponding instances in target scan, as per ground truth. \boldsymbol{\textcolor{red}{$\searrow$}} highlights differences between methods on registration and \boldsymbol{\textcolor{purple}{$\searrow$}} on matching.
    }
    \vspace{-1mm}
\label{fig:reg}
\end{figure*}

\label{chapter:experiments}
We evaluate $\more$ on its end-to-end performance on the tasks of multi-object relocalization and reconstruction~(\cref{sec:exp_end2end}), as well as on each of the three subtasks individually (\cref{sec:exp_match,sec:exp_reg,sec:exp_recon}). When evaluating end-to-end performance, we input to each subsequent task the output of the preceding one, thereby inheriting any accompanying noise and errors. When evaluating on each task independently, we provide as input the ground truth information, i.e., we provide correct instance matches to the registration task and correct registration pairs to the reconstruction task. We identify baseline methods per task and combine the best-performing ones as the end-to-end baseline. We investigate the impact of predicted instance segmentation masks as input to $\more$ and the benefit of accumulation in \cref{sec:ablation}. For more analysis on design choices, see Supp.

\subsection{Datasets}
\label{sec:exp_dataset}
In our experiments, for both $\more$ and baselines, we train on a synthetic dataset and test on both this and a real-world dataset, evaluating the generalization ability of the methods. We synthesize our own living scenes as there is no available synthetic 3D dataset of indoor scenes that exhibits long-term changes. All data will be made available.

\paragraph{\emph{FlyingShape}.} We synthesize the \emph{FlyingShape} dataset using a ShapeNet~\cite{chang2015shapenet} \textit{subset}, containing 7 categories: chair, table, sofa, pillow, bench, couch, and trash can. The training-validation split follows ShapeNet~\cite{chang2015shapenet}. We randomly sample objects from the \textit{subset}'s test set and compose 100 unique 3D scenes from them as our \textbf{test set}. We assign random poses to objects while ensuring they touch the scene's ground. To emulate long-term dynamics, we introduce random changes to all object poses and sequentially generate five temporal scans per scene. See Supp. for more details.

\paragraph{\emph{3RScan}~\cite{wald2019rio}.} \emph{3RScan} is a real-world dataset for benchmarking object instance relocalization, consisting of 1428 RGB-D sequences of 478 indoor scenes that include rescans of them. It provides annotated instance segmentation, associations, and transformations between temporal scans. 
We use the validation set of \emph{3RScan} for evaluation\footnote{The ground truth information of the \emph{3RScan} test set is hidden on a private server that is no longer maintained.}. To evaluate our comprehensive tasks, we extend \emph{3RScan} to an instance matching and reconstruction benchmark.

\subsection{Evaluation Metrics}
\label{sec:exp_metrics}
For \textbf{instance matching}, we get inspiration from the evaluation of image feature matching~\cite{sarlin2020superglue,lowe2004distinctive} and calculate the instance-level matching recall, which measures the proportion of correct matches. 

We also calculate the scene-level matching recall and use as thresholds 25\%, 50\%, 75\%, and 100\% to denote the minimum acceptable ratio of number of correct matches over the total number of matches between two temporal scans.
As \textbf{point cloud registration} is a standardized task, we follow \cite{huang2021predator,qin2022geometric,gojcic2019perfect} and report registration recall (RR), median rotation error (MedRE), transformation error (RMSE), and median Chamfer Distance (MedCD). The rotation error threshold for registration recall is $5\degree$ for \emph{FlyingShape}, and $10\degree$ for \emph{3RScan} due to its low accuracy in ground truth annotations.

For \textbf{instance reconstruction}, we follow \cite{peng2020convolutional,park2019deepsdf,mescheder2019occupancy} and report Chamfer Distance and volumetric IoU. Moreover, we introduce signed distance function (SDF) recall to describe the successful ratio of reconstruction at instance level. 

\vspace{-4mm}
\paragraph{End-to-end Metrics.}
To assess the end-to-end performance on multi-object relocalization and reconstruction, we propose two joint metrics, namely \textit{MR} recall and \textit{MRR} recall. \textit{MR} stands for the end-to-end recall of relocalization
\begin{equation}
    \mathrm{P}(R_1, M) = \mathrm{P}(R_1 | M)\mathrm{P}(M).
\end{equation}
Here $M$ and $R_1$ denote the event of an instance being correctly matched and registered, respectively. $\mathrm{P}(\cdot)$ denotes the probability of an event to happen. We pass the output of instance matching to registration and calculate registration recall (RR) as \textit{MR} recall, to include both the errors in matching and registration.
Similarly, \textit{MRR} is formulated as 

\begin{equation}
    \mathrm{P}(M, R_1, R_2) = \mathrm{P}(R_2 | R_1, M) \mathrm{P}(R_1 | M) \mathrm{P}(M),
\end{equation}
where $R_2$ denotes the event of an instance being correctly reconstructed. Here, we pass the predicted matches and the resulting registration to the reconstruction task and use the SDF recall as \textit{MRR}. As such, the performance of all three tasks is evaluated in a single metric. 

We train our model and all the baselines on the training set of the ShapeNet~\cite{xie2015deepshape} \textit{subset}, and evaluate on \emph{FlyingShape} and \emph{3RScan}. In the following quantitative evaluations, we highlight results being \colorbox{tabfirst}{best} and \colorbox{tabsecond}{second-best}.

\begin{table}[t]
    \vspace{-2mm}
    \setlength{\tabcolsep}{12pt}
    \renewcommand{\arraystretch}{1.2}
	\centering
	\resizebox{\columnwidth}{!}{
   \begin{tabular}{l|cc|cc}
    \toprule
     & \multicolumn{2}{c|}{\emph{\textbf{FlyingShape}}} &  \multicolumn{2}{c}{\emph{\textbf{3RScan}}}\\
     \midrule
       \textbf{Method} & \multicolumn{1}{c}{\textit{\textbf{MR}} \textbf{Recall} $\uparrow$} & \textit{\textbf{MRR}} \textbf{Recall} $\uparrow$ &\multicolumn{1}{c}{\textit{\textbf{MR}} \textbf{Recall} $\uparrow$} & \textit{\textbf{MRR}} \textbf{Recall} $\uparrow$\\
    \midrule
   
    Baseline$^\dag$ &67.32 &54.30 & 44.02& 30.77 \\
    Ours &\cellcolor{tabfirst}74.39 &\cellcolor{tabfirst} 62.00 & \cellcolor{tabfirst} 49.07 &  \cellcolor{tabfirst}40.74 \\
    \bottomrule
    \end{tabular}
 }
    \caption{\textbf{End-to-end performance.} \textit{MR} evaluates joint matching and registration, while \textit{MRR} measures all tasks.}
    \label{tab:end2end}
    \vspace{-1mm}
\end{table}
\begin{table}[t]
    \setlength{\tabcolsep}{12pt}
    \renewcommand{\arraystretch}{1.4}
	\centering
	\resizebox{\linewidth}{!}{
   \begin{tabular}{l|c|ccc}
    \toprule
    
       &  \multicolumn{1}{c|}{\textbf{Instance-level}} & \multicolumn{3}{c}{\textbf{Scene-level Recall}} \\  
     \midrule\arrayrulecolor{black} 
      \multicolumn{1}{l|}{\textbf{Method}} & \textbf{Recall} $\uparrow$ & \textbf{R@50}\%$\uparrow$ & \textbf{R@75}\%$\uparrow$  & \textbf{R@100}\%$\uparrow$\\
    \midrule
    MendNet~\cite{duggal2022mending} & 83.69 & \cellcolor{tabsecond}96.75 & 68.25 & 60.75  \\
    VN-DGCNN$_{\mathrm{cls}}$~\cite{deng2021vector} & 61.37 & 73.50 & 32.25 & 27.75\\

    VN-ONet$_{\mathrm{recon}}$~\cite{deng2021vector}& \cellcolor{tabsecond}86.63 & 96.00 & \cellcolor{tabsecond}74.50 & \cellcolor{tabsecond}67.75\\
    
    Ours & \cellcolor{tabfirst} 88.75 & \cellcolor{tabfirst}97.50 & \cellcolor{tabfirst}78.00 & \cellcolor{tabfirst}72.00\\
    \bottomrule
    \end{tabular}
    }
	\caption{\textbf{Instance matching results on \emph{FlyingShape}.} }
	\label{tab:shape_matching_flying}
 \vspace{-1mm}
\end{table}
\begin{table}[t]
    \setlength{\tabcolsep}{8pt}
    \renewcommand{\arraystretch}{1.4}
	\centering
	\resizebox{\linewidth}{!}{
   \begin{tabular}{l|ccc|ccc}
    \toprule
       &  \multicolumn{3}{c|}{\textbf{Instance-level Recall} $\uparrow$} & \multicolumn{3}{c}{\textbf{Scene-level Recall} $\uparrow$} \\  
     \midrule\arrayrulecolor{black} 
      \multicolumn{1}{l|}{\textbf{Method}} & \textbf{Static} & \textbf{Dynamic} & \textbf{All} & \textbf{R@25}\% & \textbf{R@50}\% & \textbf{R@75}\%  \\
    \midrule
    MendNet~\cite{duggal2022mending} & \cellcolor{tabsecond}60.32 & 63.76 & 62.20 & 80.68 & 64.77 & 37.50\\
    
    VN-DGCNN$_{\mathrm{cls}}$~\cite{deng2021vector} & 43.39 & 49.34&46.65 & 72.32&53.41 & 29.55 \\

    VN-ONet$_{\mathrm{recon}}$~\cite{deng2021vector}& 56.08& \cellcolor{tabsecond}72.05 & \cellcolor{tabsecond}64.83&  \cellcolor{tabsecond}86.36 & \cellcolor{tabsecond}71.59 & \cellcolor{tabsecond}44.32\\
    
    Ours &  \cellcolor{tabfirst}60.32 & \cellcolor{tabfirst}87.50& \cellcolor{tabfirst}71.77&\cellcolor{tabfirst}87.50 & \cellcolor{tabfirst}78.41 & \cellcolor{tabfirst}50.00\\
    \bottomrule
    \end{tabular}
    }
	\caption{\textbf{Instance matching results on \emph{3RScan}~\cite{wald2019rio}.}}
	\label{tab:shape_matching_3rscan}
 \vspace{-1mm}
\end{table}

\subsection{End-to-end Performance}
\label{sec:exp_end2end}
We combine the best-performing baseline methods in each task as the \textbf{end-to-end baseline} (Baseline$^\dag$), which comprises VN-ONet~\cite{deng2021vector}, GeoTransformer~\cite{qin2022geometric}, and ConvONet~\cite{peng2020convolutional}. As shown in \cref{tab:end2end}, our method consistently outperforms the baseline method across all metrics. Notably, there is a similar performance decrease for both methods when comparing numbers on \emph{FlyingShape} to those on \emph{3RScan}, as anticipated due to the inherent domain gap. Also, there is an increased gap between our method and the baseline in end-to-end evaluation vs. the results on individual tasks (\cref{sec:exp_match} to \ref{sec:exp_recon}), for both relocalization\footnote{3RScan~\cite{wald2019rio} provides a baseline for instance relocalization. We do not compare with it because it takes as input TSDF patches, not point clouds, and it is not reproducible due to missing codebase and inadequate details.}  and reconstruction. This disparity is attributed to our unified approach, utilizing a single network and representation that retains shape and pose information. In contrast, the combined baseline lacks coherence between tasks, employing three distinct networks and representations. Qualitative results on \emph{3RScan}~\cite{wald2019rio} are in \cref{fig:end_to_end} for end-to-end performance (MRR) and in \cref{fig:reg} for relocalization (MR).
\begin{figure}[t]
\vspace{-2mm}
    \centering
    \includegraphics[width=\linewidth]{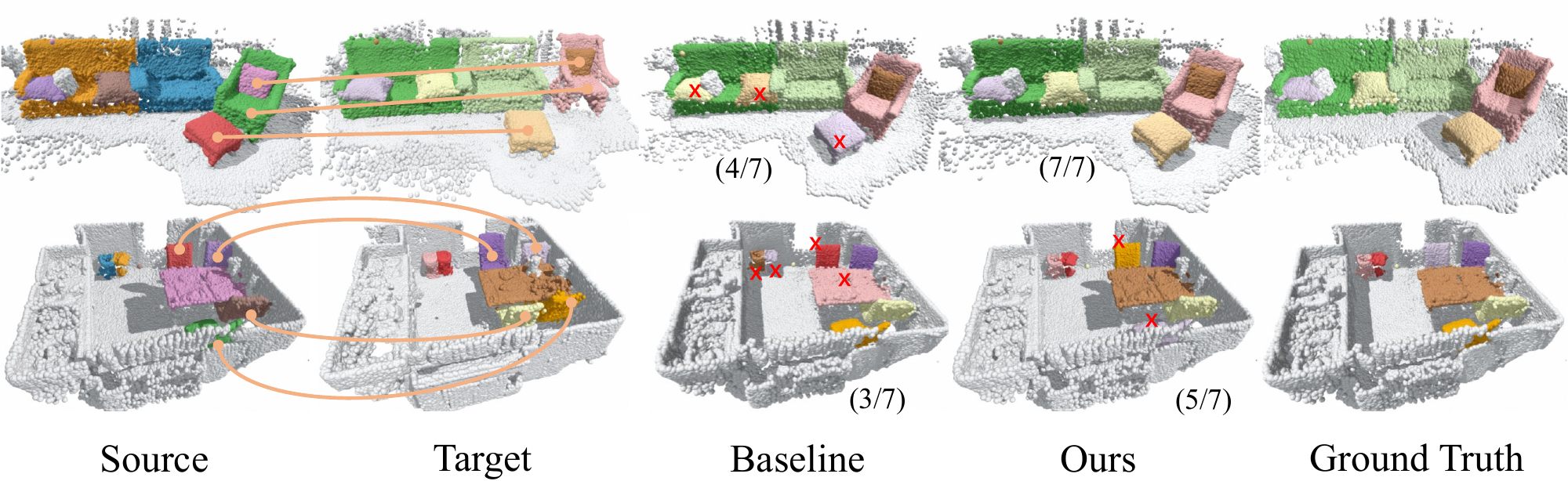}
    \vspace{-15pt}
    \caption{ \textbf{Multi-object matching on \emph{3RScan}~\cite{wald2019rio}.} We repaint the instances in the source scan using the same colors as those of matched instances in the target scan. \textcolor{red}{X} denotes the wrongly matched instances. \colorbox{tabsecond}{Curves} depict the associations of moving objects. \textit{(5/7)} denoting 5 correct matches out of 7 pairs in the scene.}
    \label{fig:shape_matching}
\end{figure}

\subsection{Instance Matching}
\label{sec:exp_match}
We compare $\more$ with three baselines, namely MendNet~\cite{duggal2022mending}, VN-ONet~\cite{deng2021vector,mescheder2019occupancy},
and VN-DGCNN~\cite{wang2019dgcnn,deng2021vector}, of which the first two are point cloud reconstruction networks, and the last is a point cloud classification network. We present the results in \cref{tab:shape_matching_flying} and \cref{tab:shape_matching_3rscan} for \textit{FlyingShape} and \textit{3RScan}, respectively. $\more$ outperforms the baseline methods on all metrics. This can be attributed to the representation power of its embeddings: the encoder can output expressive global invariant features to handle large pose variations, and the equivariant features model the high-frequency details of the input point cloud. In \cref{fig:shape_matching}, we showcase that $\more$ can handle in-category object matching by capturing minor geometric variations.
\begin{table}[t]
    \setlength{\tabcolsep}{8pt}
    \renewcommand{\arraystretch}{1.2}
	\centering
	\resizebox{\columnwidth}{!}{
   \begin{tabular}{clcccc}
    \toprule

        \textbf{Dataset} & \textbf{Method} & \selectfont \textbf{RR} $\uparrow$ & \textbf{MedRE} $\downarrow$ &  \textbf{RMSE} $\downarrow$  & \textbf{MedCD}  $\downarrow$ \\
      \midrule
      \multirow{5}{*}{\emph{FlyingShape}} &RPMNet~\cite{yew2020rpm}&  23.17 & 2.37 & 31.77 & 0.0249   \\
      &FreeReg~\cite{zhu2022correspondence}& 47.50 & 2.44 & 33.84 & 0.0760   \\
      &GeoTransformer~\cite{qin2022geometric}&  77.67 & 1.36 & \cellcolor{tabfirst}16.66 & 0.0271  \\
      \arrayrulecolor{black!10}\cmidrule{2-6}\arrayrulecolor{black} 
      &Ours w/o optim& \cellcolor{tabsecond}83.00 & \cellcolor{tabsecond}0.86 & 20.83 & \cellcolor{tabsecond}0.0171 \\
      &Ours full & \cellcolor{tabfirst}83.83 & \cellcolor{tabfirst}0.74 & \cellcolor{tabsecond}18.47 & \cellcolor{tabfirst}0.0168 \\

   \hline

      \multirow{5}{*}{\emph{3RScan}~\cite{wald2019rio}} &  RPMNet~\cite{yew2020rpm}& 9.40 & 3.78& 15.91 & 0.0248 \\
      &FreeReg~\cite{zhu2022correspondence}& 26.06 & 5.76& 11.05 & 0.0082   \\
      &GeoTransformer~\cite{qin2022geometric}& 51.71 & \cellcolor{tabfirst} 3.46 & 6.51 &0.0141 \\
      \arrayrulecolor{black!10}\cmidrule{2-6}\arrayrulecolor{black} 
      &Ours w/o optim &  \cellcolor{tabsecond}58.12 & 3.77 & \cellcolor{tabsecond} 5.49  & \cellcolor{tabsecond}0.0032 \\
      &Ours full & \cellcolor{tabfirst} 61.11 & \cellcolor{tabsecond} 3.77 & \cellcolor{tabfirst} 4.74  & \cellcolor{tabfirst}0.0030 \\
      \bottomrule
    \end{tabular}
    }
    \caption{\textbf{Point cloud registration results on both datasets.}}
    \label{tab:registration}
    \vspace{-1mm}
\end{table}

\subsection{Point Cloud Registration}
\label{sec:exp_reg}
\begin{figure}[t]
\vspace{-2mm}
    \centering
    \includegraphics[width=.95\linewidth]{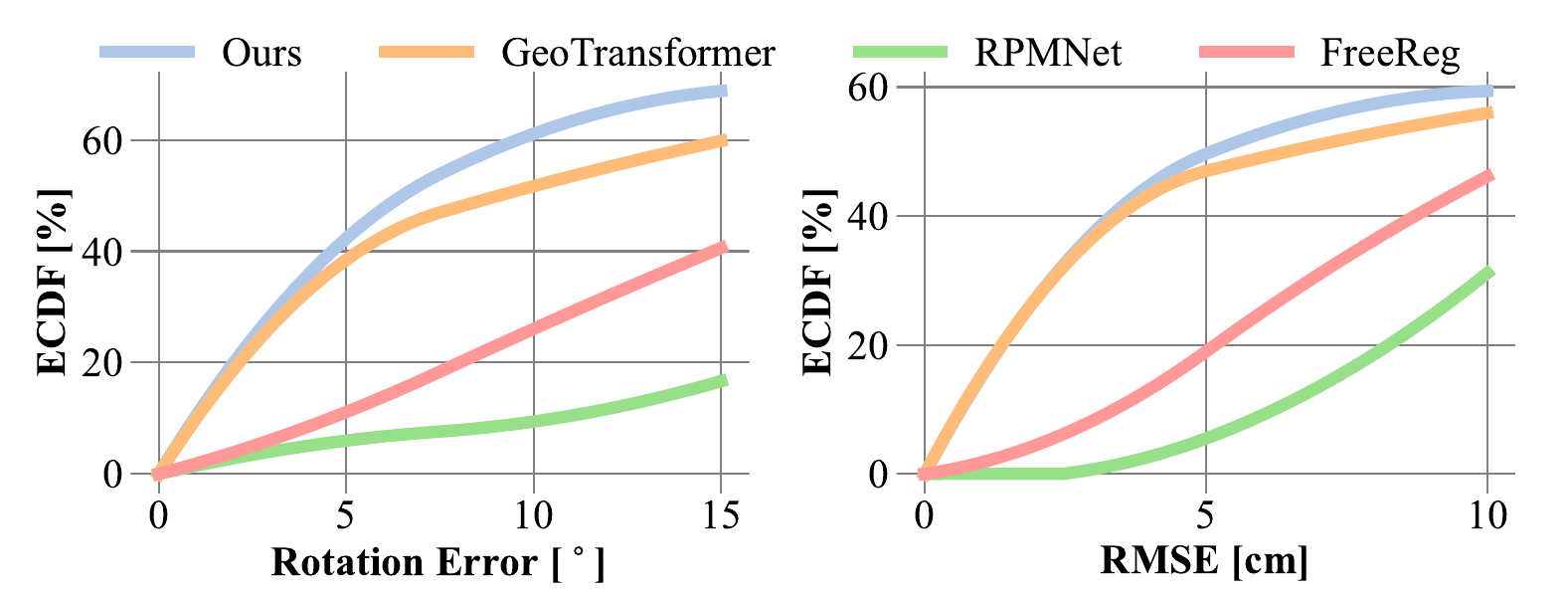}
    \vspace{-5pt}
    \caption{\textbf{ECDF curves of registration results on \emph{3RScan}~\cite{wald2019rio}}. }
\label{fig:reg_ecdf}
\end{figure}

We compare $\more$ with three baselines: RPMNet~\cite{yew2020rpm}, which is a learning-based method that only targets object-level registration; FreeReg~\cite{zhu2022correspondence}, which uses equivariant embeddings to solve rotation; and GeoTransformer~\cite{qin2022geometric}, which is the state-of-the-art method on 3DMatch~\cite{zeng20173dmatch} and 3DLoMatch~\cite{huang2021predator} datasets. Results on \textit{FlyingShape} and \textit{3RScan} are in \cref{tab:registration}. \cite{zhu2022correspondence} can only provide coarse registration and does not work well under large partiality changes. In contrast to \cite{yew2020rpm} and \cite{qin2022geometric}, $\more$ does not rely on discrete point-wise correspondences but represents the geometry with continuous signed distance field and aligns the point cloud with the field via optimization. Our analysis is further corroborated by the highest Empirical Cumulative Distribution Function (ECDF) curve of $\more$ in \cref{fig:reg_ecdf}. 

\begin{table}[t]
    \setlength{\tabcolsep}{4pt}
    \renewcommand{\arraystretch}{1.3}
	\centering
	\resizebox{\columnwidth}{!}{
   \begin{tabular}{l|ccc|cc}
    \toprule
    \textbf{Dataset} & \multicolumn{3}{c|}{\textbf{\emph{FlyingShape}}} & \multicolumn{2}{c}{\textbf{\emph{3RScan}}} \\
     \midrule
      \textbf{Method}  & \bm{$L1$}\textbf{-Chamfer}\bm{$_\mathrm{2}$} $\downarrow$ & \textbf{IoU} $\uparrow$&\textbf{SDF Rec.}$\uparrow$ & \bm{$L1$}-\textbf{Chamfer}\bm{$_\mathrm{1}$} $\downarrow$ & \textbf{SDF Rec. $\uparrow$}\\
    \midrule
   
    MendNet~\cite{duggal2022mending} & 25.27 & 47.79 & 6.17& 17.73 & 20.99   \\
    VN-ONet~\cite{deng2021vector} & 8.55 & 34.47 & 65.00& 10.65 & 51.91   \\
    ConvONet~\cite{peng2020convolutional} &6.64 & 36.99 & \cellcolor{tabsecond}80.67&  \cellcolor{tabsecond} 7.61 & \cellcolor{tabfirst} 64.89  \\
     \arrayrulecolor{black!10}\midrule\arrayrulecolor{black} 
     Ours w/o optim  & \cellcolor{tabsecond}6.27 & \cellcolor{tabsecond}49.98& 78.00& 9.28 & 56.87 \\
    Ours full & \cellcolor{tabfirst}6.11 & \cellcolor{tabfirst}66.73 & \cellcolor{tabfirst}83.33& \cellcolor{tabfirst} 6.16 & \cellcolor{tabsecond} 64.12 \\
    \bottomrule
    \end{tabular}
     }
	\caption{\textbf{Instance reconstruction results}. $L1$-Chamfer $\times10^{-3}$.}
	\label{tab:recon}
\end{table}

\subsection{Instance Reconstruction}
\label{sec:exp_recon}
We compare \more{} with MendNet~\cite{duggal2022mending}, VN-ONet~\cite{deng2021vector} and ConvONet~\cite{peng2020convolutional}; results are in \cref{tab:recon}. We report the 2-way chamfer on \emph{FlyingShape} and only the 1-way chamfer on \emph{3RScan} as it only provides incomplete object meshes (non-watertight). With joint optimization, our full method surpasses the baselines on most metrics across the two datasets. Without, it is on par with ConvONet on \emph{FlyingShape}. This demonstrates the adaptation power of our optimization algorithm on noisy and randomly oriented point clouds. In contrast to baseline methods that only perform surface reconstruction, our pose graph $\mathbf{G}$ and shared embedding $\mathbf{F}$ enable optimization message passing and fusion between accumulated point clouds, improving both registration and reconstruction performance (\cf \cref{tab:registration} and \cref{tab:recon}).

\vspace{0mm}
\section{Ablation Study}
\label{sec:ablation}
\paragraph{Predicted instance segmentation.}
Noisy and incomplete instance segmentation masks from Mask3D~\cite{schult2023mask3d} are provided to \more{}. Results in \cref{tab:ablation_mask3D} across all tasks and combinations, when compared to GT masks, show our method outperforming the combined baseline. More importantly, MR and MRR recall is substantially lower for the baseline, despite a similar matching and registration recall for both methods. Also, the drop of the baseline in MR and MRR recall between GT and Mask3D~\cite{schult2023mask3d} is $\approx 33\%$, vs $\approx 18\%$ in ours. These findings demonstrate the efficacy of \more{}'s shared representation across tasks and joint optimization, even in noisy settings.        

\begin{table}[t]
    \vspace{-2mm}
    \setlength{\tabcolsep}{12pt}
    \renewcommand{\arraystretch}{1.2}
	\centering
	\resizebox{\columnwidth}{!}{
   \begin{tabular}{lcccccc}
    \toprule
       \textbf{Method} & \textbf{Ins. Seg.} &\multicolumn{1}{c}{\textbf{Mat. Rec.} $\uparrow$}  & \textbf{Reg. Rec.} $\uparrow$   & \textit{\textbf{MR}} \textbf{Rec.} $\uparrow$ & \textit{\textbf{MRR}} \textbf{Rec.} $\uparrow$\\
    \midrule
   
    Baseline$^\dag$ & \multirow{2}{*}{GT} & 64.83 & 51.71& 44.02 & 30.77\\
    Ours & &\cellcolor{tabfirst}71.77& \cellcolor{tabfirst}61.11 & \cellcolor{tabfirst}49.07& \cellcolor{tabfirst}40.74 \\
    \midrule
    Baseline$^\dag$ & \multirow{2}{*}{Mask3D~\cite{schult2023mask3d}} & 43.43  & 47.74  & 27.86 & 20.89\\
    Ours & & \cellcolor{tabfirst}45.76 & \cellcolor{tabfirst}51.27 & \cellcolor{tabfirst}40.14 & \cellcolor{tabfirst}33.80\\
    \bottomrule
    \end{tabular}
    }
	\caption{\textbf{Results with Mask3D~\cite{schult2023mask3d} on \emph{3RScan}~\cite{wald2019rio}.}}
    \label{tab:ablation_mask3D}
    \vspace{-1mm}
\end{table}

\begin{figure}[t]
    \centering
    \includegraphics[width=\linewidth]{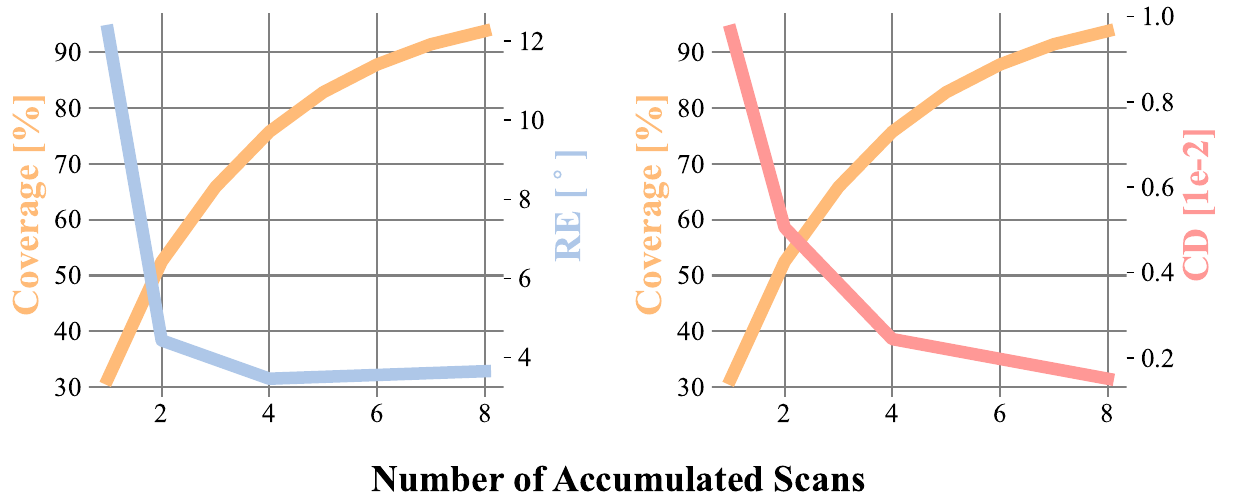}
    \caption{\textbf{Ablation study on point cloud accumulation.} The change of point cloud coverage, rotation error and chamfer distance w.r.t the number of accumulated scans. 
    }
    \label{fig:benefit}
\end{figure}

\paragraph{Benefit of accumulation.} The results in \cref{tab:registration} are computed for pairs of point clouds, \ie, between two points in time. Here, we experiment with increasing the number of multi-temporal scans used for accumulation (\cf \cref{fig:benefit}) and report the performance of registration (RE) and reconstruction (CD) on \emph{FlyingShape}, to showcase an increasing geometric accuracy and completeness over time. 
We see significant improvement on both metrics in the range of no accumulation (one point cloud) to four point clouds, after which the performance starts to saturate. The saturation is explained when compared to the coverage ratio of accumulated point cloud w.r.t. the complete shape. By the 4th scan, completeness is close to 75\%, hence any additional scan will affect less registration and reconstruction.

\section{Conclusion}
\label{sec:conclusion}
We propose $\more$, a novel approach to parse long-term dynamic scenes (living scenes) involving three consecutive tasks. $\more$ solves the three tasks by flexible adaptation of equivariant embeddings and a joint optimization that enables multi-temporal accumulation. Our approach exhibits superior performance across both synthetic and real-world datasets. It empowers the cumulative comprehension of 3D assets in the scene. Future research directions involve addressing challenges posed by the presence of elastic deformations and multiple identical objects in the scene and comprehending large-scale spatiotemporal changes \cite{sun2023NSS}.

\clearpage
{
\small
\bibliographystyle{ieeenat_fullname}
\bibliography{main}
}

\setcounter{section}{0}
\renewcommand\thesection{\Alph{section}}
\clearpage
\setcounter{page}{1}
\maketitlesupplementary
In this supplementary material, we provide: 
\begin{enumerate}
\item A video that explains our method and shows our experimental results (\cref{supp:video}).
\item Additional ablation study results (\cref{sec:ablation}) and detailed per category results (\cref{sec:cate_performance}).
\item Implementation details on the network architecture (\cref{{supp:architecture}}) and algorithm for registration and joint optimization (\cref{supp:algo}).
\item Details on data processing for training and evaluation (\cref{sec:data_processing}).
\item Description of evaluation metrics (\cref{supp:metrics}).
\item Additional visualizations on multi-object relocalization and reconstruction 
(\cref{sec:more_visual}).
\item A discussion of the limitations of our method (\cref{supp:discussion}).
\end{enumerate}

\section{Video}
\label{supp:video}
See the provided supplementary video for a summarized description of the method and results on creating living scenes.

\section{Additional Experimental Results}
In this section, we present additional experimental results. Specifically, we further validate our design choices with three ablation studies (\cref{sec:ablation}). In~\cref{sec:cate_performance}, we provide detailed per category results. We provide more qualitative results in \cref{sec:more_visual}.

\subsection{Ablation Studies}
\label{sec:ablation}
We ablate $\more$ to justify our design choices for instance matching, joint optimization, and network architecture.
\paragraph{Embeddings for matching.} We compare the performance of different combinations of embeddings to compute the score matrix used for instance matching on \emph{3RScan}~\cite{wald2019rio}. The results are tabulated in \cref{tab:ablation_matching}. When compared individually, the invariant embedding $\Fin$ performs consistently better than the equivariant $\Feqv$, signifying it plays a more important role in instance matching. This is not a surprise, since the shape details can be more decisive than pose, especially when there is a changing environment. However, when using both embeddings, the results are further boosted. Hence, we use both embeddings in \more{}.
\begin{table}[t]
    \setlength{\tabcolsep}{2pt}
    \renewcommand{\arraystretch}{1.4}
    \centering	
    \resizebox{\columnwidth}{!}{
    \begin{tabular}{cc|cccccc}
    \toprule
      \multicolumn{2}{c|}{\textbf{Matching Setting}}  &  \multicolumn{6}{c}{\textbf{Evaluation Metrics}} \\
      \midrule
    \multirow{2}{*}[-.4em]{$\mathbf{E}$ ($\Feqv$)}  & 
    \multirow{2}{*}[-.4em]{$\mathbf{\Lambda}$ ($\Fin$)} &

    \multicolumn{3}{c}{\textbf{Instance-level Recall} $\uparrow$} & \multicolumn{3}{c}{\fontsize{10}{12}\selectfont \textbf{Scene-level Recall} $\uparrow$} \\ 
    \cmidrule(lr){3-5}
    \cmidrule(rr){6-8}
    && \textbf{Static}  & \textbf{Dynamic}  & \textbf{All}   & \textbf{@25\%} & \textbf{@50\%}  &\textbf{@75\%} \\
    
    \midrule
     
      & \cmark  & 58.20  &  \cellcolor{tabsecond}78.60 & \cellcolor{tabsecond}69.38 & \cellcolor{tabfirst}89.77 & \cellcolor{tabfirst}79.55 &  \cellcolor{tabsecond}45.45 \\
      \cmark & &\cellcolor{tabsecond}58.20  & 73.80 & 66.75 & 85.23 & 72.73 & 43.18    \\
      \cmark& \cmark  & \cellcolor{tabfirst} 60.32 & \cellcolor{tabfirst}87.50& \cellcolor{tabfirst}71.77 & \cellcolor{tabsecond}87.50& \cellcolor{tabsecond}78.41 & \cellcolor{tabfirst}50.00  \\

    \bottomrule
    \end{tabular}
    } 
    \caption{\textbf{Ablation study of instance matching on \emph{3RScan}~\cite{wald2019rio}}. \cmark\ denotes using the embedding for instance matching.} 
    \vspace{-1mm}
    \label{tab:ablation_matching}
\end{table}
\begin{table}[t]
    \setlength{\tabcolsep}{4pt}
    \renewcommand{\arraystretch}{1.2}
    \centering	
    \resizebox{\columnwidth}{!}{
    \begin{tabular}{cccc|cccc}
    \toprule
      \multicolumn{4}{c|}{\textbf{Optimization Setting}}  &  \multicolumn{4}{c}{\textbf{Evaluation Metrics}} \\
      \midrule
    \multirow{2}{*}[-.4em]{$\mathbf{G}$} & 
    \multirow{2}{*}[-.4em]{$\Feqv$}  & 
    \multirow{2}{*}[-.4em]{$\Fin$} & 
    \multirow{2}{*}[-.4em]{$\Fcentroid$} & 
    
    \multicolumn{2}{c}{\textbf{RE [$\degree$]}} & \multicolumn{2}{c}{\fontsize{9}{12}\selectfont \textbf{$L1$-Chamfer} [$\times10^{-3}$] } \\ 
    \cmidrule(lr){5-6}
    \cmidrule(rr){7-8}
    &&&& \textbf{Mean} $\downarrow$ & \textbf{Median} $\downarrow$ & \textbf{Mean} $\downarrow$ & \textbf{Median} $\downarrow$ \\
    
    \midrule
    \xmark & \xmark & \xmark& \xmark  & 11.85  & 5.28  & 5.37  & 2.31   \\
    \cmark & \xmark & \xmark& \xmark  & \cellcolor{tabsecond}7.88 & 3.45          & 5.38          & 2.30   \\
    \cmark & \xmark & \xmark& \cmark  & \cellcolor{tabfirst}7.81 & 3.41          & 5.67          & 2.30    \\
    \cmark & \cmark & \xmark& \cmark  & 8.38 & \cellcolor{tabsecond}1.81& \cellcolor{tabsecond}3.95 &  \cellcolor{tabsecond}2.06   \\
    \cmark & \cmark & \cmark& \cmark  & 8.70 & \cellcolor{tabfirst}1.71 & \cellcolor{tabfirst}3.81 & \cellcolor{tabfirst}1.86   \\

    \bottomrule
    \end{tabular}
    } 
    \caption{\textbf{Ablation study on joint optimization}. \xmark\ denotes that the parameters are fixed during optimization and \cmark denotes learnable parameters.} 
    \vspace{-1mm}
    \label{tab:ablation_optim}
\end{table}
\begin{table}[t]
    \setlength{\tabcolsep}{10pt}
    \renewcommand{\arraystretch}{1.2}
	\centering
	\resizebox{\columnwidth}{!}{
   \begin{tabular}{lcccc}
    \toprule
      \textbf{Parameter}  &$\mathbf{F_{inv}}$ & $\mathbf{F_{t}}$ & $\mathbf{F_{eqv}}$ & $\mathbf{G}$ \\
    \midrule
     Step Size [$\times10^{-5}$] & 1 & 1 & 10 & 5\\
    \bottomrule
    \end{tabular}
    }
	\caption{\textbf{Step sizes of different parameters during joint optimization}. }
	\label{tab:lr}
\end{table}

\paragraph{Parameters to optimize.} To find the best set of parameters for the joint optimization, we experiment with different optimization settings on \emph{FlyingShape}. \cref{tab:ablation_optim} shows the optimization setting on the left and the corresponding performance on the right. The best performance is achieved when we optimize all four parameters. The optimization on equivariant embeddings significantly reduces the rotation error. Pure optimization on the pose graph gives the best rotation error but provides poor performance on reconstruction. This also shows that a single forward pass from the VN-encoder cannot provide accurate pose and shape information and verifies the importance of jointly optimizing pose and shape together. We use different learning rates (step size) during optimization for each parameter (\cref{tab:lr}). We mainly optimize $\mathbf{F_{eqv}}$ and $\mathbf{G}_0$ and only apply small adjustments to $\mathbf{F_{inv}}$ and $\mathbf{F_{t}}$.  
\begin{figure*}[t]
    \centering
    \includegraphics[width=.8\textwidth]{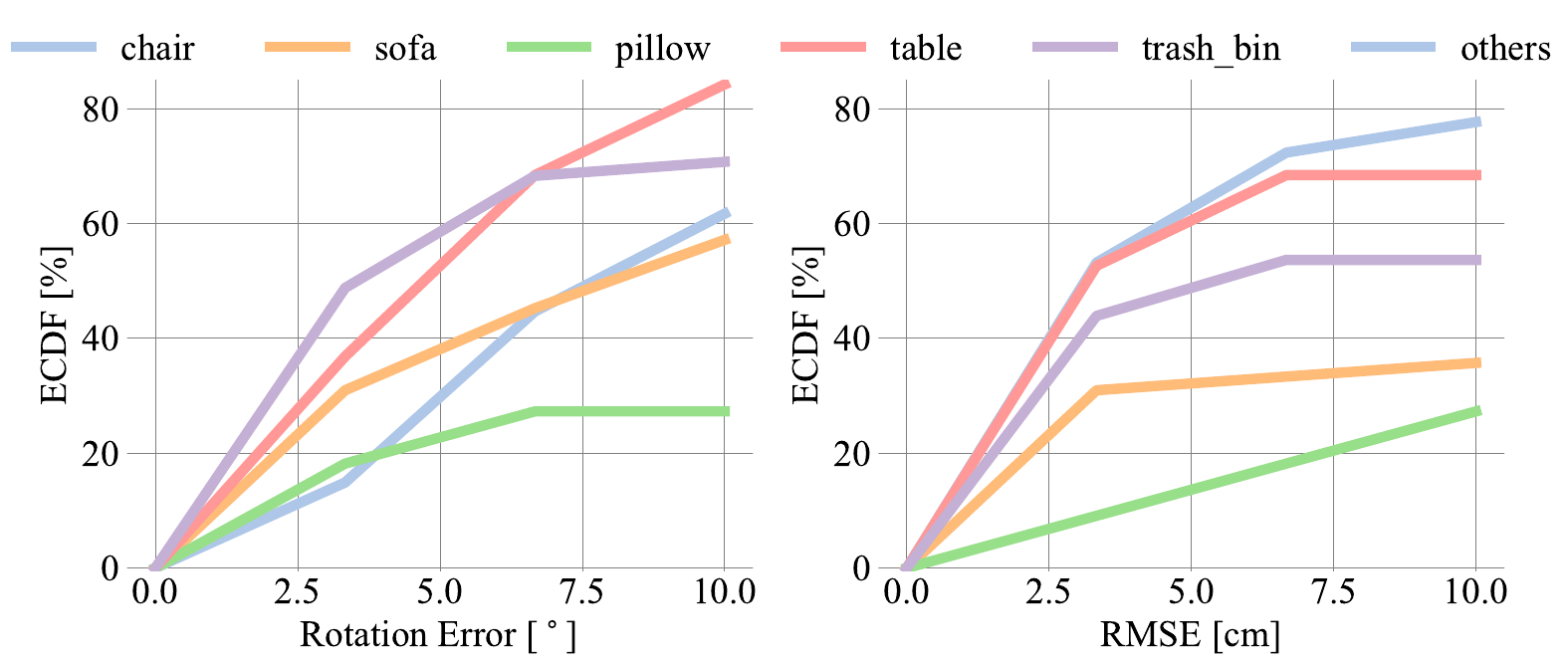}
    \caption{ \textbf{Per category registration results on \emph{3RScan}~\cite{wald2019rio}.} We report ECDF curves of rotation error (RE) and transformation error (RMSE).} 
\label{fig:per_cate_performance}
\end{figure*}

\paragraph{Decoding strategy.} 
We explore four decoding strategies using equivariant $\Feqv$ and invariant $\Fin$ embeddings on \emph{FlyingShape}. The DeepSDF decoder takes as input the positional embedding of the coordinates to query and the global latent code of the shape representation. For a query point $\mathbf{p} = (x, y, z)$, there are two ways to compute the positional embedding:
\begin{itemize}
\item Direct concatenation (inv.): $\mathrm{cat}[\Fin, \mathbf{p}]$
\item Concatenating with the inner product (inner.) of equivariant embeddings and query point: $\mathrm{cat}[\Fin, \langle \Feqv, \mathbf{p}\rangle]$
\end{itemize}
where $ \mathrm{cat}[\cdot, \cdot]$ denotes feature-wise concatenation

As decoding architectures, we consider DeepSDF~\cite{park2019deepsdf} and a multi-layer perceptron (MLP)~\cite{haykin1998mlp}. DeepSDF differs from regular MLPs, as shown in \cref{fig:decoder}, by adding a skip connection of the query code. We evaluate the performance of combining each decoding strategy with each of the decoder architectures and present the results in \cref{tab:decoding}. Inner.+DeepSDF shows the best performance. The inner product can map the query coordinates to a high dimensional space with more positional information as provided by the skip connection in DeepSDF~\cite{park2019deepsdf}.

\subsection{Category-level Performance}
\label{sec:cate_performance}
We report the registration performance of each category on \emph{3RScan}~\cite{wald2019rio} in \cref{fig:per_cate_performance}. The category of `\textit{table}' has the overall best performance and we believe this is due to the distinct geometric features of tables and their relatively large size in the scene. The category of pillows has the worst performance, which is not surprising since it is not strictly rigid and has multiple symmetrical axes. 
\begin{table}[t]
    \setlength{\tabcolsep}{4pt}
    \renewcommand{\arraystretch}{1.2}
	\centering
	\resizebox{\columnwidth}{!}{
   \begin{tabular}{l|l|cccc}
    \toprule
     \textbf{PE} & \textbf{Decoder}   & \textbf{$L1$-Chamfer} $\downarrow$ & \textbf{$L1$-UNI} $\downarrow$ & \textbf{$L1$-NSS} $\downarrow$ & \textbf{IoU} $\uparrow$ \\
        \midrule

Inv. & MLP       & 7.04            & 0.037            & 0.021                 & 0.325          \\
Inv. & DeepSDF   & 7.34            & 0.041            & 0.019                 & 0.343           \\
Inner. & MLP     & \cellcolor{tabsecond}3.65  & \cellcolor{tabsecond}0.029  & \cellcolor{tabsecond}0.016  & \cellcolor{tabsecond}0.436           \\
Inner. & DeepSDF & \cellcolor{tabfirst}3.58 & \cellcolor{tabfirst}0.027 & \cellcolor{tabfirst}0.015 & \cellcolor{tabfirst}0.454  \\
    \bottomrule
    \end{tabular}
    }
	\caption{\textbf{Ablation study on decoding strategies}. Positional Encoding (PE). Inner product (Inner.) and invariant feaures (Inv.). $L1$-Chamfer [$\times10^{-3}$], UNI - Uniform SDF Sampling, NSS - Near Surface Sampling.}
	\label{tab:decoding}
 \vspace{-2mm}
\end{table}

\section{Network Architecture}
\label{supp:architecture}
In this section, we provide the implementation details of the encoder-decoder network used in the main paper.

\begin{figure*}[t]
    \centering
    \includegraphics[width=\textwidth]{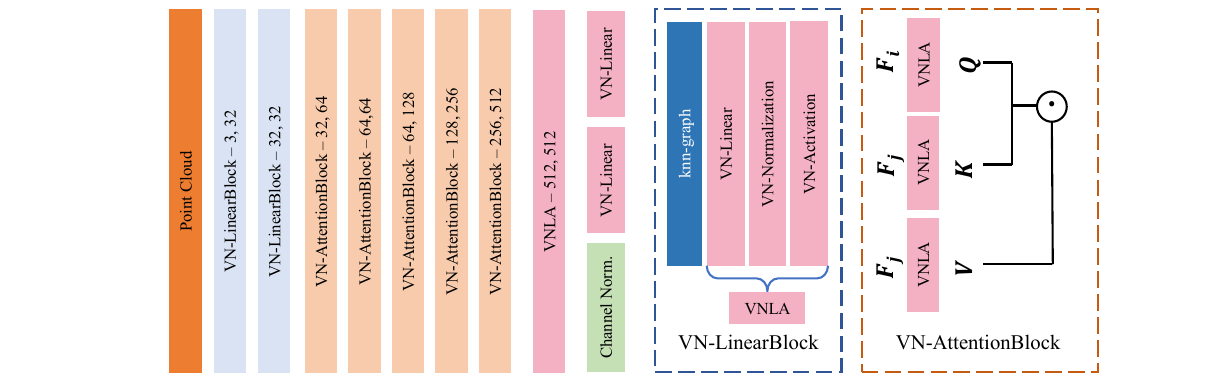}
    \caption{\textbf{VN-Encoder Architecture~\cite{deng2021vector,assaad2022vn_transformer}.}  On the left is the overall architecture from the input point cloud to the intermediate embeddings, which are subsequently fed to the DeepSDF decoder. $\odot$ denotes dividing the feature into multiple heads and computing score matrix $S$ from $Q$ and $K$, and message passing between $S$ and $V$, lastly to the final output feature of the encoder. On the right, we show two important blocks of the encoder: VN-LinearBlock and VN-AttentionBlock. }
\label{fig:arch_encoder}
\end{figure*}

\subsection{Encoder}
Here, we follow recent progress~\cite{Lei2023EFEM,assaad2022vn_transformer} in equivariant networks in vector neurons (VN)~\cite{deng2021vector} and present the graphic illustration in \cref{fig:arch_encoder}. The encoder takes as input 3D point cloud coordinates and processes them through 2 VN-Linear blocks and 5 VN-Attention blocks. The VN-Linear block consists of computing the k-nearest neighbor graph feature~\cite{wang2019dgcnn}, and the VNLA block, which comprises three layers: VN-Linear, VN-normalization, and VN-activation~\cite{deng2021vector}. 
The VN-Attention block~\cite{assaad2022vn_transformer} utilizes three VNLA blocks to compute the \textbf{\textit{K}}, \textbf{\textit{Q}}, and \textbf{\textit{V}}, respectively, and applies the attention operation~\cite{vaswani2017attention} and message passing. The dimensions of the features for the 7 blocks are: 32, 32, 64, 64, 128, 256, and 512. This means that the intermediate features are down-sampled after the 2nd (VN-Linear), 4th (VN-Attention), and 5th (VN-Attention) blocks. The final global feature of dimension 512 is then passed to compute the final output $\mathbf{F} = (\Fin \in \mathbb{R}^{256}, \Feqv \in \mathbb{R}^{3\times 256}, \Fscale  \in \mathbb{R}_{+}, \Fcentroid  \in \mathbb{R}^{3})$. Here, $\Fin, \Feqv$, and $\Fcentroid$ are computed by three VN-Linear prediction heads and $\Fscale$ from channel-wise normalization.

\subsection{Decoder}
\begin{figure}[t]
    \centering
    \includegraphics[width=.9\textwidth]{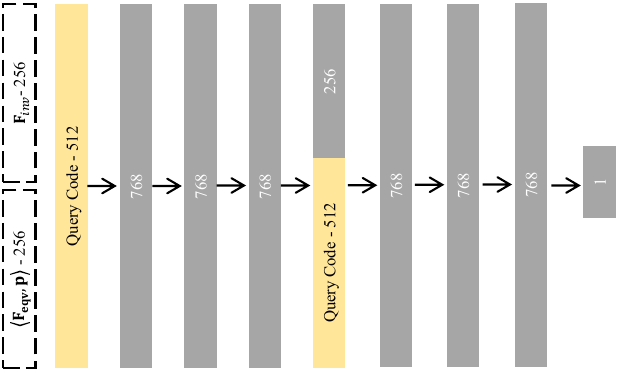}
    \caption{\textbf{DeepSDF Decoder.} $\mathbf{\rightarrow}$ denotes residual connection. The query code is re-concatenated to the intermediate feature at the 4th layer. }
    
\label{fig:decoder}
\end{figure}

The architecture of the decoder is presented in \cref{fig:decoder}. The number of fully connected layers is 8, the same as in the original DeepSDF~\cite{park2019deepsdf}. In \more{}, the feature dimension is increased from 256 in DeepSDF to 768 for two reasons: (1) the dimension of the positional embedding is 256 and not 3; and (2) to increase the expressivity of the decoder when trained on multiple categories (category-agnostic). Same as in \cite{park2019deepsdf}, the query code is computed by concatenating the shape code, \ie, $\Fin$, and the positional embedding $\langle \Feqv, (\mathbf{p}-\Fcentroid)/\Fscale \rangle$, which is further re-concatenated to the intermediate feature of the 4th layer (skip connection). The skip connection~\cite{park2019deepsdf} can provide better performance and a more regularized reconstruction (\cf Inner.+MLP vs. Inner.+DeepSDF in \cref{tab:decoding}).

\begin{figure}[t]
    \centering
    \includegraphics[width=1.0\textwidth]{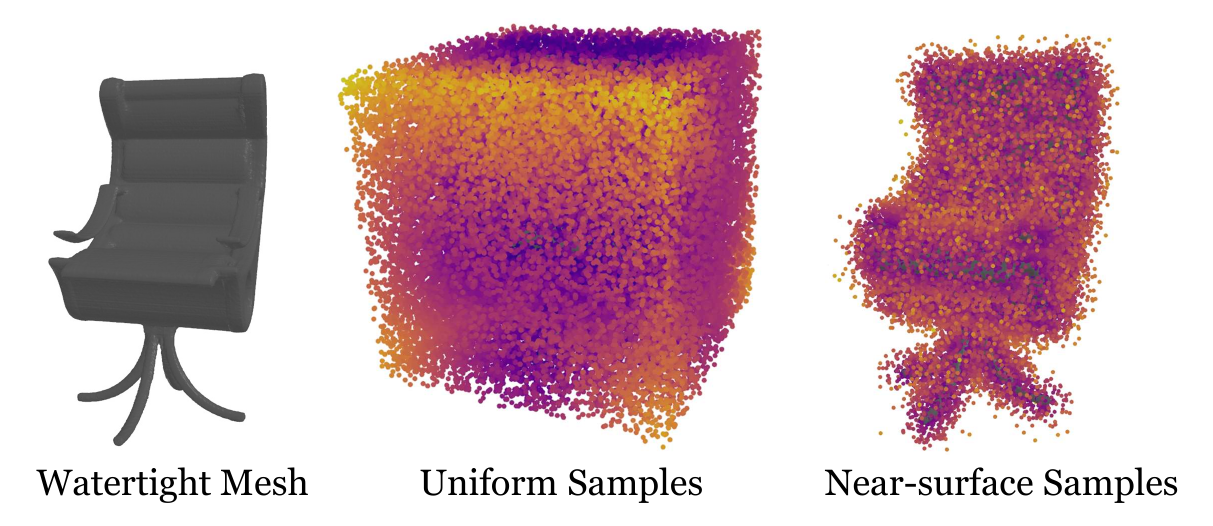}
    \caption{\textbf{Illustration of SDF samples for training.} Left: a watertight mesh. Middle: uniform SDF samples. Right: near-surface SDF samples.}
\label{fig:sdf_sampling}
\end{figure}

\section{Training Details}
\label{supp:loss}
We follow EFEM~\cite{Lei2023EFEM} and DISN~\cite{xu2019disn} and train the network using SDF loss with regularization terms.
\paragraph{SDF Loss.} The SDF samples are generated from a unit cube around the object. 50\% are sampled near the surface and 50\% uniformly in the space (\cf \cref{fig:sdf_sampling}). We encourage the network to learn more local details with: 
\begin{equation}
\mathcal{L}_{\text {SDF }}=\frac{\lambda_{\text {near }} \sum_{x \in \mathcal{Q}_{\text {near }}} \mathcal{L}_\mathrm{recon}(x)+\lambda_{\text {far }} \sum_{x \in \mathcal{Q}_{\text {far }}} \mathcal{L}_\mathrm{recon}(x)}{\left|\mathcal{Q}_{\text {near }}\right|+\left|\mathcal{Q}_{\text {far }}\right|}.
\end{equation}
where $\mathcal{Q}_{\text {near}}$ denotes the set of samples with $|\mathbf{SDF}|<0.1$ and $\mathcal{Q}_{\text {far }}$ the set of samples with $|\mathbf{SDF}|\geq0.1$. The weights of the two sets of samples are $\lambda_{\text {near }} = 1.0$ and $\lambda_{\text {far }} = 0.5$.
\paragraph{Regularization~\cite{Lei2023EFEM}.} Two loss terms are used to regularize the training of scale and centroid prediction: 
\begin{equation}
\mathcal{L}_{\text {scale }}=\left|1.0-\Fscale\right|
\end{equation}
and
\begin{equation}
\mathcal{L}_{\text {center }}=\left\|\Fcentroid\right\|_2.
\end{equation}
The regularization forces the scale to be one and the center of the shape to be at the origin because ShapeNet~\cite{chang2015shapenet} meshes are placed at their canonical space. The centroid regularization provides signals to correct the center of gravity of a partial point cloud to its actual center in the canonical space during training.

The final loss is computed as follows:
\begin{equation}
\mathcal{L}=\omega_{\text {SDF }} \mathcal{L}_{\text {SDF }}+\omega_{\text {center }} \mathcal{L}_{\text {center }}+\omega_{\text {scale }} \mathcal{L}_{\text {scale }},
\end{equation}
We follow~\cite{Lei2023EFEM} and set $\omega_{\text {SDF }}=1.0$, $\omega_{\text {center }} =0.2$, and $\omega_{\text {scale }}=0.01$.

\section{Algorithms}
\label{supp:algo}
\paragraph{Registration.} We provide the details of our proposed registration in \cref{ag:pairwise_reg}. We set $\Obj^{t_1}$ as source and $\Obj^{t_2}$ as target in the registration and use the encoder $\Phi$ and decoder $\Psi$ for initialization and optimization, respectively.  $\mathrm{Kabsch(\cdot,\cdot)}$ denotes Kabsch algorithm~\cite{kabsch1976solution}. $\mathbf{P}_\mathrm{i}$ denotes the source point cloud transformed by the estimated $(\mathbf{R}_\mathrm{i}, \mathbf{t}_\mathrm{i})$ at $\mathrm{i}^{th}$ iteration. $J$ denotes the analytical Jacobians of $(\mathbf{R}_\mathrm{i}, \mathbf{t}_\mathrm{i})$. $\nabla$ denotes the gradient of parameters.

\paragraph{Joint Optimization.} We provide the algorithmic details of our proposed joint optimization in \cref{ag:optim}. We take accumulated point clouds of each instance $\{\Obj^{t} | t \in \{1,..\mathrm{T}\}\}$ as input. $\mathrm{REG()}$ denotes the registration algorithm in \cref{ag:pairwise_reg} and we use it to initialize the pose graph $\mathbf{G}$. We compute the loss $\loss_{joint}$ of the accumulated point cloud, which is the sum of the SDF loss $\loss_{sdf}$ and the regularization loss $\loss_{z}$.

\RestyleAlgo{ruled}
\SetKwComment{Comment}{/* }{ */}

\begin{algorithm}[t]
\caption{Registration}\label{alg:joint_optim}
\KwIn{\\$\Obj^{t_1}$(source), $\Obj^{t_2}$(target), $\Phi$(encoder), $\Psi$(decoder)}
\Comment{Initialization}
$\mathbf{F}^{t_1}, \mathbf{F}^{t_2}$ $ \gets \Phi(\Obj^{t_1}),\Phi(\Obj^{t_2})$ \\
$\mathbf{R}, \mathbf{t} \gets \mathrm{Kabsch}(\mathbf{F}^{t_1},\mathbf{F}^{t_2})$ \\

$\eta \gets 10^{-3}$: step size,
${K} \gets 200$: number of steps\;

\Comment{Iterative Update}
\For{$\mathrm{i} = 0,...,K $}{
 $\mathbf{P}_\mathrm{i} \gets \mathbf{R}_\mathrm{i}\Obj^{t_1} +\mathbf{t}_\mathrm{i}$ \\
 $\mathbf{F}_{q}^{t_2} \gets \mathrm{cat}[\Fin^{t_2}, \langle\Feqv^{t_2}, (\mathbf{P}_\mathrm{i}-\Fcentroid^{t_2}) / \Fscale^{t_2}\rangle]$ \\
 $\loss \gets \loss_{reg}(\Obj^{t_2}, \Obj^{t_2})$ \\
 $J(\mathbf{R}_\mathrm{i}, \mathbf{t}_\mathrm{i}) = \loss_{reg}\left(\Obj^{t_2}, \Obj^{t_2} \right)$ \\
 $\mathbf{R}_\mathrm{i+1} \gets \mathbf{R}_\mathrm{i} - \eta \cdot \nabla_\mathbf{R} J( \mathbf{R}_\mathrm{i}, \mathbf{t}_\mathrm{i})$ \\
 $\mathbf{t}_\mathrm{i+1} \gets \mathbf{t}_\mathrm{i} - \eta \cdot \nabla_\mathbf{t} J( \mathbf{R}_\mathrm{i}, \mathbf{t}_\mathrm{i})$
 } 

\Comment{Terminate Iteration}
\KwOut{$\mathbf{R}, \mathbf{t} \gets \argmin_{\mathbf{R}, \mathbf{t}}\mathcal{L}$ }
\label{ag:pairwise_reg}
\end{algorithm}

\RestyleAlgo{ruled}
\SetKwComment{Comment}{/* }{ */}

\begin{algorithm}[t]
\caption{Joint Optimization}\label{alg:joint_optim}
\KwData{$\{\Obj^{t} | t \in \{0,1,..\mathrm{T-1}\}\}$}
\Comment{Initialization}
$\mathbf{F}^1,\mathbf{F}^1,...\mathbf{F}^{T}  \gets \Phi(\Obj^0)$\; 
\For{$t = 0,...,T-1$}{
 $\mathbf{T}_t \gets \mathrm{REG}(\Feqv^t, \Feqv^{t+1})$\;}
 $\mathbf{G} \gets \{ {T}_t\}_{t=1}^{T}$\;
 $\mathbf{F^*} \gets \argmin_{\mathbf{F}} \{ \loss_{sdf}(\mathbf{F}^t) \}_{t=1}^{T}$\;
$\epsilon$: learning rate,
$I \gets 200$: number of steps\;

\Comment{Iteration}
\While{$\mathrm{i} < I$}{
  $\mathcal{L}_{joint} \gets 0$\;
  \For{$t = 0,...,T-1$}{
 $\mathcal{L}_{joint} \textrm{ += } \mathcal{L}_{sdf}(\Obj^t) + \mathcal{L}_{z}(\mathbf{F}_\mathrm{i})$\ ;} 
  \Comment{Update using Adam}
  $[\mathbf{F}_\mathrm{i}, \mathbf{G}_\mathrm{i}] \gets \mathrm{AdamUpdate}(\mathcal{L}_{joint}, \epsilon)$ \;
  $\mathrm{i} \gets \mathrm{i}+1$;
}
\Comment{Terminate Iteration}
\KwOut{$[\mathbf{F}, \mathbf{G}] = \argmin_{\mathbf{F}, \mathbf{G}}\mathcal{L}_{joint}$}
\label{ag:optim}
\end{algorithm}

\section{Data Processing}
\label{sec:data_processing}
\subsection{Training Data}
The network is trained under the supervision of Signed Distance Fields (SDFs). In order to generate the training samples, we compute the SDF for every shape in the training set of the ShapeNet~\cite{chang2015shapenet} \textit{subset}. We partly follow the processing pipelines of \cite{mescheder2019occupancy,Lei2023EFEM}, which include three steps: \emph{(i)} making the mesh watertight, \emph{(ii)} generating point clouds from partial mesh
renderings, and \emph{(iii)} sampling SDFs. 

\paragraph{Making the mesh watertight.} Watertight meshes usually describe meshes consisting of one closed surface. This means that they do not contain holes and have a clearly defined interior~\cite{stutz2023watertight}. The CAD models in ShapeNet are non-watertight. We use MeshFusion\footnote[1]{ \href{https://github.com/davidstutz/mesh-fusion}{https://github.com/davidstutz/mesh-fusion}}~\cite{Stutz2018meshfusion} to process raw meshes to watertight. 

\paragraph{Generating point clouds from mesh renderings.} To mimic partial observations as in real-world datasets, we render depth maps of meshes from multiple views. First, we construct a sphere around the mesh, fully covering it and placing it exactly at the center. We uniformly sample 24 points on the sphere as focal points of the depth camera. To ensure that the majority of the shape is within the field of view, we place the principal point on the line connecting the focal point and the center of the sphere by solving the camera orientation $\mathbf{R}$:
\begin{equation}
    z\begin{bmatrix} w/2\\h/2\\1\end{bmatrix} = \mathbf{K}[\mathbf{R}|\mathbf{T}] \begin{bmatrix} X\\Y\\Z\\1\end{bmatrix},
\end{equation}
where $\mathbf{K}$ denotes camera intrinsics and $[\mathbf{R}|\mathbf{T}]$ transformation from world frame to camera frame. $w$ and $h$ are the dimensions of the image and $[X,Y,Z,1]^T$ are the coordinates of the focal point in the world frame. $\mathbf{R}$ is the unknown to solve. We render depth maps at the 24 sampled positions for every mesh using OpenGL~\cite{woo1999opengl} and back-project them to 3D space as point clouds. 

\paragraph{Sampling SDFs~\cite{Lei2023EFEM}.} We sample SDF values around meshes. We adopt two sampling strategies: uniform sampling and near-surface sampling (\cf \cref{fig:sdf_sampling}). Uniform sampling captures the global structure of the shape and near-surface sampling captures high-frequency (detailed) signals. For each mesh, we sample $10^5$ SDF samples, 50\% uniform and 50\% near the surface.

\subsection{FlyingShape Dataset}
\begin{figure*}[t]
    \centering
    \includegraphics[width=0.9\linewidth]{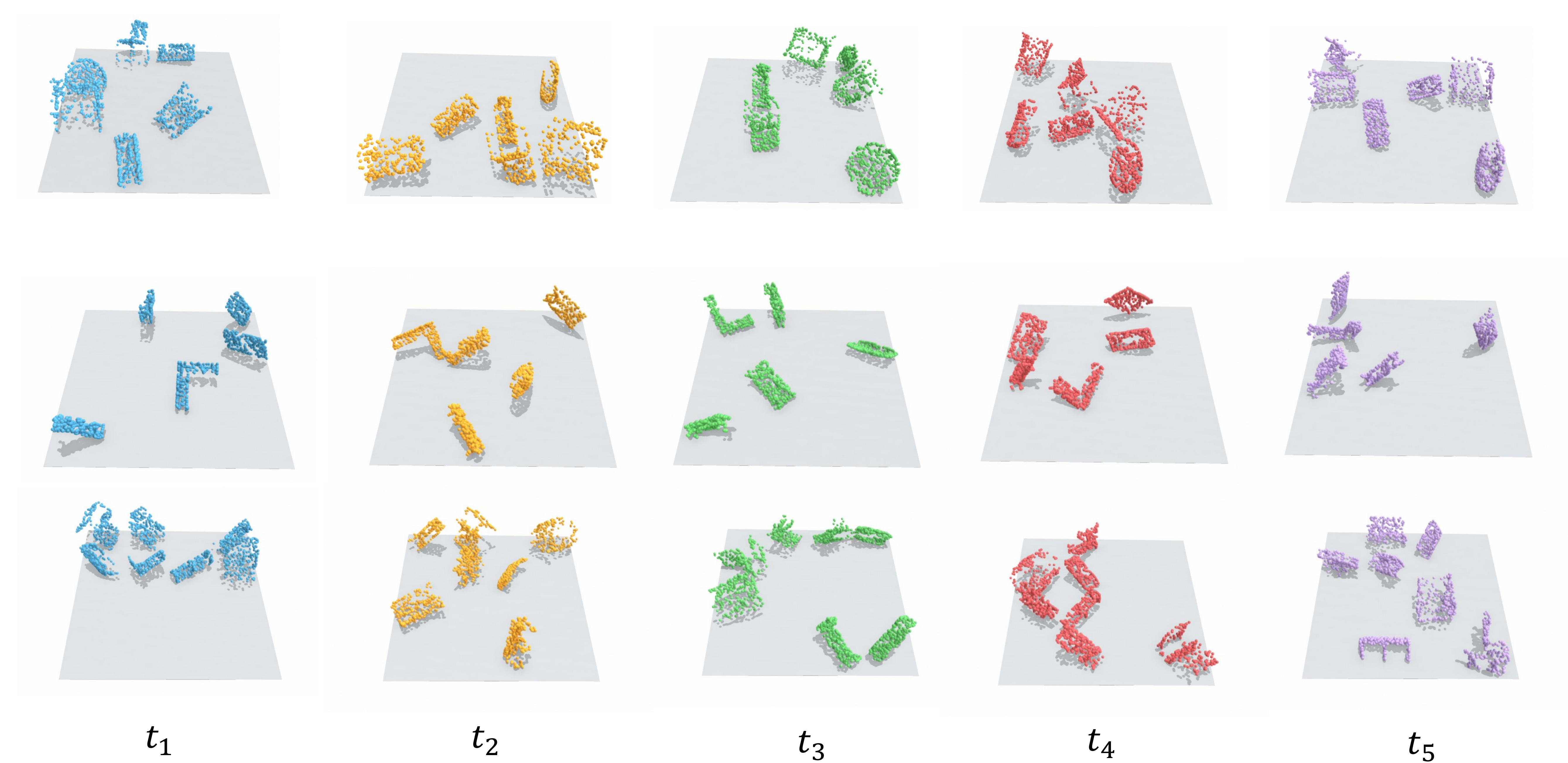}
    \caption{\textbf{Samples of changing scenes from \emph{FlyingShape}.} Each scene has five temporal scans, shown in different colors.}
\label{fig:flyingdataset}
\end{figure*}

We synthesize the \emph{FlyingShape} dataset based on the \textit{subset} of ShapeNet's test set. It consists of scenes that have been captured repeatedly at irregular intervals (temporal scans), in between which consisting objects have been moved.
The number of objects in each scene ranges from four to eight. As the number of objects per category in the \textit{subset} are different, we balance their frequency when randomly drawing samples. To replicate the partial completeness of a scene in real captures, we generate the scan of each scene by rendering  depth images from random viewpoints on the upper hemisphere within the scene and un-projecting the depth pixels into the 3D space. We combine the point clouds resulting from three rendered views. We create 100 scenes in total, with each scene containing five temporal scans. We generate annotations on instance segmentation, associations, and transformations directly from the ground truth. Sample scenes are shown in \cref{fig:flyingdataset}. 

\subsection{\emph{3RScan} Dataset}
The \emph{3RScan} dataset~\cite{wald2019rio} provides raw RGB-D scans with known poses. As our method reasons on point cloud, we downsample the RGB-D frames to reduce the point cloud density and back-project them to obtain the raw scan as our input. We use the semantic maps of 23 categories: 

\textit{armchair, bed, bench, chair, coffee table, computer desk, couch, couch table, cushion, desk, dining chair, dining table, footstool, ottoman, pillow, rocking chair, round table, side table, sofa, sofa chair, stand, table } and \textit{trash can}.

We set the other semantic labels as background and do not process them. We use the official toolbox of \cite{wald2019rio} and generate the instance masks for our generated point cloud from the dataset's semantic mesh. As \cite{wald2019rio} only provides the mesh reconstruction on the level of a scene, we use the instance masks to extract the individual mesh (vertices and faces) of each instance in the scene as ground truth.

\section{Evaluation Metrics}
\label{supp:metrics}
In this section, we provide the formulas for all the evaluation metrics we used in the main paper, in the order of the three sequential tasks.

\subsection{Instance Matching}
\textbf{Instance-level matching recall.} This metric measures the fraction of correctly matched instance pairs over ground truth instance pairs within the entire dataset.
\begin{equation}
    \textrm{Instance Recall} = \frac{ \#\textrm{ correct matches}}{\#\textrm{ total matches}}
\end{equation}

\noindent \textbf{Scene-level matching recall.} To understand the performance on a scene level, since each scene has multiple instances, we compute instance recall only for instance pairs that exist in the scene and check what fraction of scenes have recall $> \tau$. In the main paper, we set $\tau$ to four values 25\%, 50\%, 75\%, and 100\%, of which the first three are used for \emph{3RScan}~\cite{wald2019rio} and the last three are used for \emph{FlyingShape}, as \emph{3RScan} is a real-world dataset and is more challenging than \emph{FlyingShape}.

\subsection{Point Cloud Registration.}
\paragraph{Rotation error (RE).} It measures the geodesic distance between two rotations:
\begin{equation}
\mathrm{RE}=\arccos \left(\frac{\operatorname{trace}\left(\mathbf{R}^\mathrm{T} \overline{\mathbf{R}}\right)-1}{2}\right),
\end{equation}
where $\mathbf{R}$ denotes the predicted rotation matrix and $\overline{\mathbf{R}}$ the ground truth. $\operatorname{trace}()$ denotes the trace of a matrix.
\paragraph{Registration recall (RR).} It is the fraction of rotation errors smaller than a threshold. We use as thresholds  
$\mathrm{RE}<5^\circ$ for \emph{FlyingShape} and  $\mathrm{RE}<10^\circ$ for \emph{3RScan}~\cite{wald2019rio}. We use a higher threshold for 3RScan because the accuracy of ground truth transformations is lower, since they were obtained using Procrustes on manually annotated 3D keypoint correspondences between two temporal scans~\cite{wald2019rio}. \emph{FlyingShape} is synthesized with no introduced errors.

\paragraph{Transformation error.} It is the root mean square error of per-point distance between point clouds transformed by prediction and ground truth poses. 
\begin{equation}
\resizebox{\hsize}{!}{$
\begin{aligned}
\mathrm{RMSE} = \sqrt{\frac{1}{M+N} \left(\sum\left\|\mathbf{T}_{\mathbf{P}}^{\mathbf{Q}}(\mathbf{p})-\overline{\mathbf{T}}_{\mathbf{P}}^{\mathbf{Q}}(\mathbf{p})\right\|_2^2 + \sum\left\|\mathbf{T}_{\mathbf{Q}}^{\mathbf{P}}(\mathbf{q})-\overline{\mathbf{T}}_{\mathbf{Q}}^{\mathbf{P}}(\mathbf{q})\right\|_2^2\right)}
\end{aligned}$}
\end{equation}
where $\mathbf{T}_{\mathbf{P}}^{\mathbf{Q}}$ denotes the transformation from $\mathbf{P}$ to $\mathbf{Q}$ and vice versa.
$M,N$ denote the number of points in $\mathbf{P},\mathbf{Q}$, respectively.

\paragraph{Chamfer distance (CD).} It measures the quality of registration. Here we follow~\cite{huang2021predator,yew2020rpm} and use the modified chamfer distance metric:
\begin{equation}
\begin{aligned}
\tilde{C D}(\mathbf{P}, \mathbf{Q})= \frac{1}{|\mathbf{P}|} \sum_{\mathbf{p} \in \mathbf{P}} \min _{\mathbf{q} \in \mathbf{Q}_{\text {raw }}}\left\|\mathbf{T}_{\mathbf{P}}^{\mathbf{Q}}(\mathbf{p})-\mathbf{q}\right\|_2^2+ \\
 \frac{1}{|\mathbf{Q}|} \sum_{\mathbf{q} \in \mathbf{Q}} \min _{\mathbf{p} \in \mathbf{P}_{\text {raw }}}\left\|\mathbf{q}-\mathbf{T}_{\mathbf{P}}^{\mathbf{Q}}(\mathbf{p})\right\|_2^2 ,
\end{aligned}
\end{equation}
where $\mathbf{P}_{\text {raw }} \in \mathbb{R}^{M \times 3}$ and $\mathbf{Q}_{\text {raw }} \in \mathbb{R}^{N \times 3}$ are \emph{raw} source and target point clouds, and $\mathbf{P} \in \mathbb{R}^{1024 \times 3}$ and $\mathbf{Q} \in \mathbb{R}^{1024 \times 3}$ are \emph{input} source and target point clouds.

\paragraph{Empirical Cumulative  Distribution Function (ECDF).} This metric  
measures the distribution of a set of values
\begin{equation}
\operatorname{ECDF}(x)=\frac{\left|\left\{o_i<x\right\}\right|}{|O|},
\end{equation}
where $O$ is a set of values (RE and RMSE in our case) and $x \in[\min \{O\}, \max \{O\}]$~\cite{huang2021predator}. We compute the ECDF curves w.r.t. rotation error and transformation error in the main paper.

\subsection{Instance Reconstruction} 
We follow the definition from \cite{mescheder2019occupancy, peng2020convolutional}. Let $\mathcal{M}_{\text {pred }}$ and $\mathcal{M}_{\mathrm{GT}}$ be the sets of all points that are inside or on the surface of the predicted and ground truth mesh, respectively.
\paragraph{Chamfer-$L1$.} We follow~\cite{peng2020convolutional} and define reconstruction accuracy and completeness:
\begin{equation}
\label{eq:ref}
\resizebox{\hsize}{!}{$
\begin{aligned}
\operatorname{Accuracy}\left(\mathcal{M}_{\text {pred }} \mid \mathcal{M}_{\mathrm{GT}}\right)  \equiv \frac{1}{\left|\partial \mathcal{M}_{\text {pred }}\right|} \int_{\partial \mathcal{M}_{\text {pred }}} \min _{\mathbf{q} \in \partial \mathcal{M}_{\mathrm{GT}}}\|\mathbf{p}-\mathbf{q}\| \mathrm{d} \mathbf{p} 
\end{aligned}$}
\end{equation}

\begin{equation}
\resizebox{\hsize}{!}{$
\begin{aligned}
    \operatorname{Complete.}\left(\mathcal{M}_{\text {pred }} \mid \mathcal{M}_{\mathrm{GT}}\right) \equiv \frac{1}{\left|\partial \mathcal{M}_{\mathrm{GT}}\right|} \int_{\partial \mathcal{M}_{\mathrm{GT}}} \min _{\mathbf{p} \in \partial \mathcal{M}_{\mathrm{pred}}}\|\mathbf{p}-\mathbf{q}\| \mathrm{d} \mathbf{q}
\end{aligned}$}
\end{equation}
Here $\partial \mathcal{M}_{\mathrm{GT}}$ and $\partial \mathcal{M}_{\mathrm{GT}}$ are the surfaces of $\mathcal{M}_{\text {pred }}$
and $\mathcal{M}_{\mathrm{GT}}$, respectively. The $\text { Chamfer-} L_1$ can be defined as below:

\begin{equation}
\resizebox{\hsize}{!}{$
\begin{aligned}
& \text { Chamfer-} L_1\left(\mathcal{M}_{\text {pred }}, \mathcal{M}_{\mathrm{GT}}\right)= \\
& \qquad \frac{1}{2}\left(\operatorname{Accuracy}\left(\mathcal{M}_{\text {pred }} \mid \mathcal{M}_{\mathrm{GT}}\right)+
\operatorname{Completeness}\left(\mathcal{M}_{\text {pred }} \mid \mathcal{M}_{\mathrm{GT}}\right)\right)
\end{aligned}$}
\end{equation}

\paragraph{Volumetric IoU.} It is equal to the intersection divided by the union of two sets:
\begin{equation}
\operatorname{IoU}\left(\mathcal{M}_{\text {pred }}, \mathcal{M}_{\mathrm{GT}}\right) \equiv \frac{\left|\mathcal{M}_{\text {pred }} \cap \mathcal{M}_{\mathrm{GT}}\right|}{\left|\mathcal{M}_{\text {pred }} \cup \mathcal{M}_{\mathrm{GT}}\right|}.
\end{equation}
We follow the implementation of ConvONet~\cite{peng2020convolutional} to compute IoU: randomly sample $100 \mathrm{K}$ points from the bounding boxes of each mesh and determine if the points lie inside or outside $\mathcal{M}_{\text {pred }}$ and $\mathcal{M}_{\mathrm{GT}}$, respectively.

\paragraph{SDF Recall.} It is designed by us to evaluate reconstruction quality when the watertight mesh of ground truth is not available:
\begin{equation}
    \textrm{SDF Recall:} = \frac{1}{K} \sum_{\mathbf{v} \in \mathbf{V}} \mathbbm{1} (\mathbf{SDF}(\mathbf{v})),
\end{equation}
where $\mathbbm{1} (\mathbf{SDF}(\mathbf{v}))=1$ if $\mathbf{SDF}(\mathbf{v})<0.05$ else $0$. $\mathbf{V}$ denotes the set of vertices of the ground truth mesh and $\mathbf{v}$ each vertex therein. We evaluate by computing the mean absolute SDF errors of vertices in the ground truth mesh (not necessarily watertight) w.r.t. the predicted mesh using library~\cite{point-cloud-utils} and calculate the ratio of vertices with an SDF error smaller than the threshold. We set the threshold of SDF values to 0.05 because the threshold for near-surface samples during training is 0.1 and we divide it by two during evaluation.

\section{Additional Qualitative Results}
\label{sec:more_visual}
\begin{figure}[h]
    \centering
    \includegraphics[width=\linewidth]{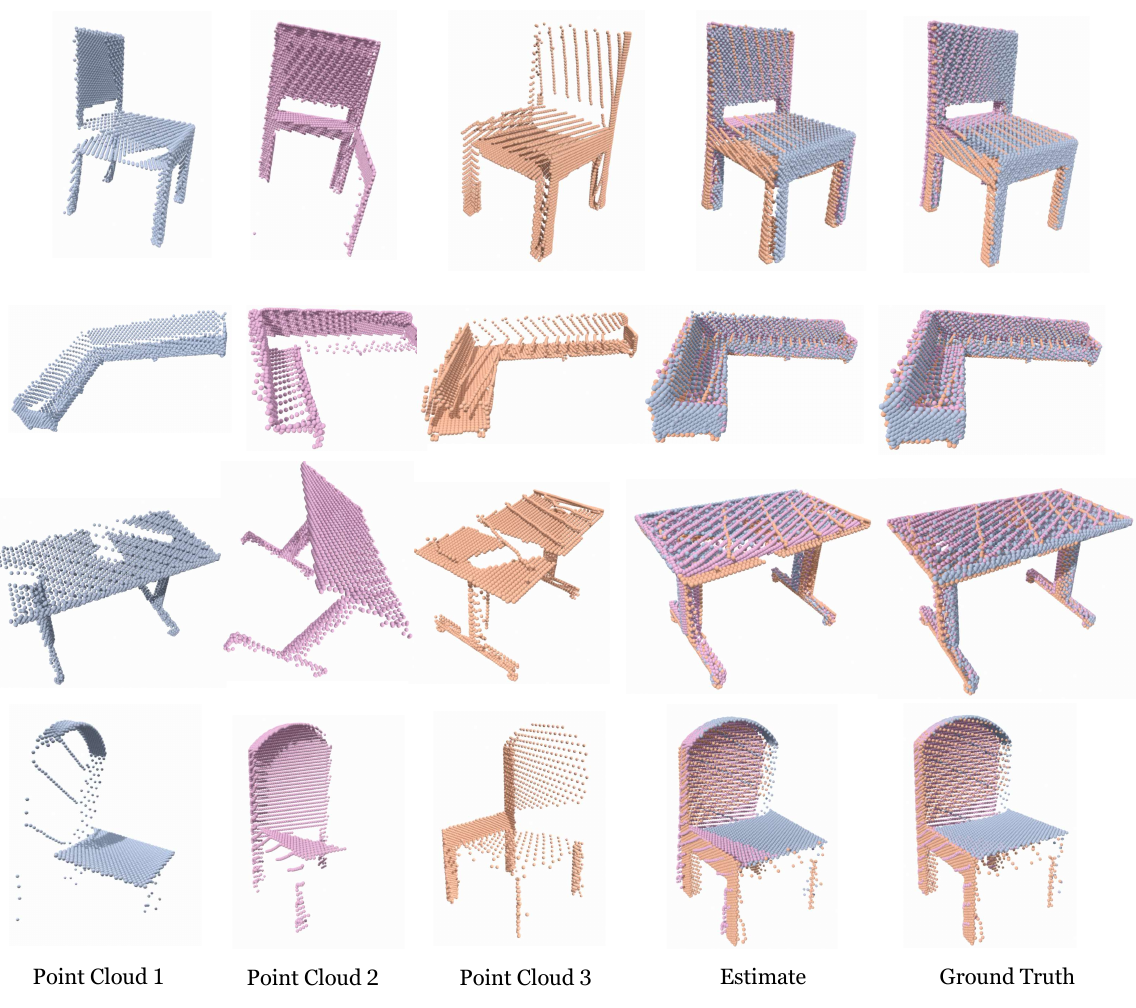}
    \caption{\textbf{Qualitative results of point cloud registration on \emph{FlyingShape}.} Left three columns are the three input point clouds. Ground truth on the right. Our method does not look for correspondences between two point clouds, \eg, the back and front of the chair in the last row: our method first tries to complete the instance surface based on partial observations and then registers the completed zero-level set. }
\label{fig:instance_reg}
\end{figure}

\begin{figure}[t]
    \centering
    \includegraphics[width=\linewidth]{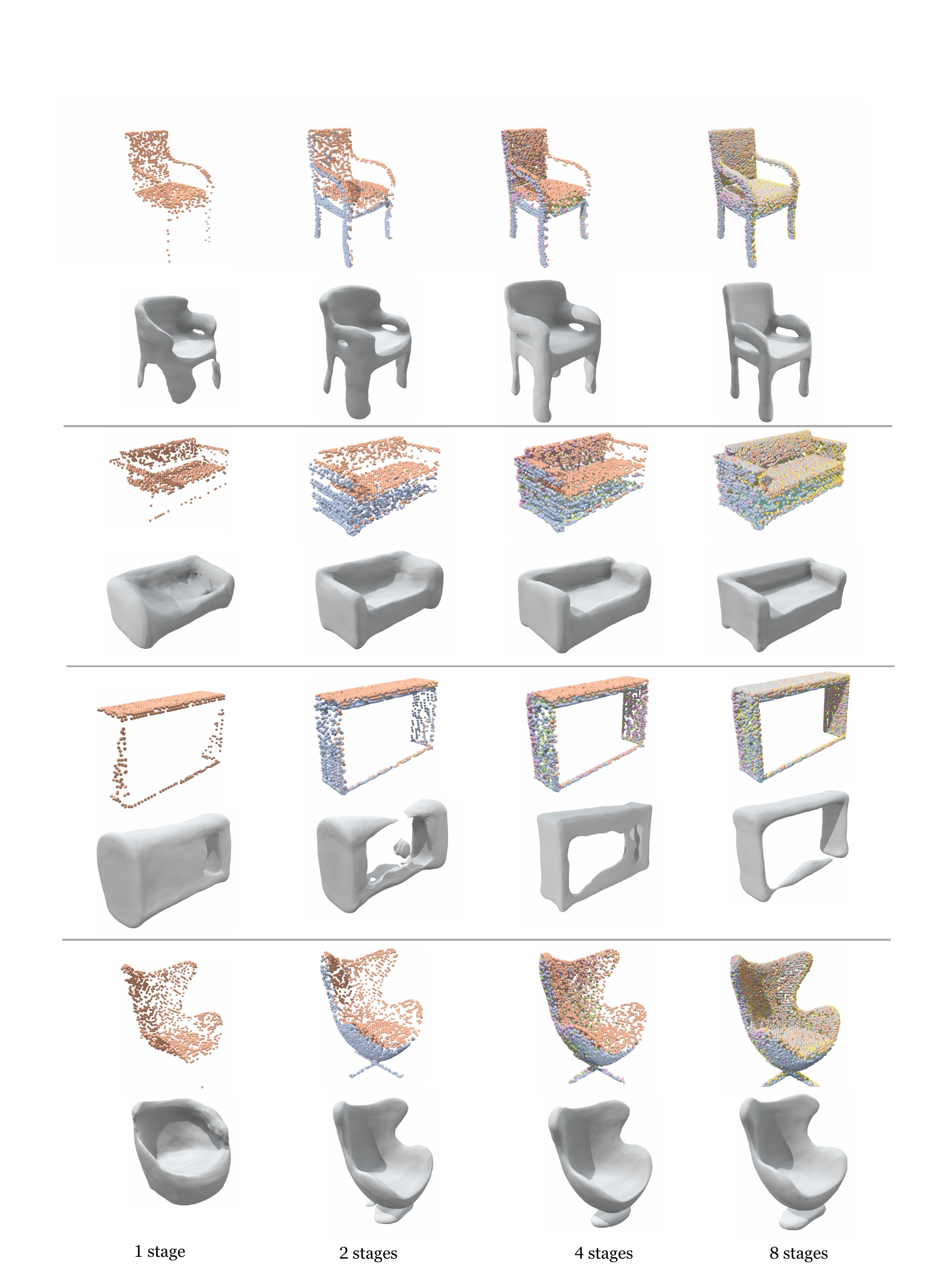}
    \caption{\textbf{Qualitative results of point cloud accumulation at instance level on \emph{FlyingShape}.} This figure is corresponding to Figure. 7 (in the main paper), showing the ever-increasing quality and completeness of reconstruction with \more{} when more data are accumulated. Each temporal point cloud is showcased in a different color. Per example, top row shows the registration and bottom row shows the reconstruction results.}
\label{fig:supp_accumulation}
\end{figure}

\begin{figure*}[t]
    \centering
    \includegraphics[width=\linewidth]{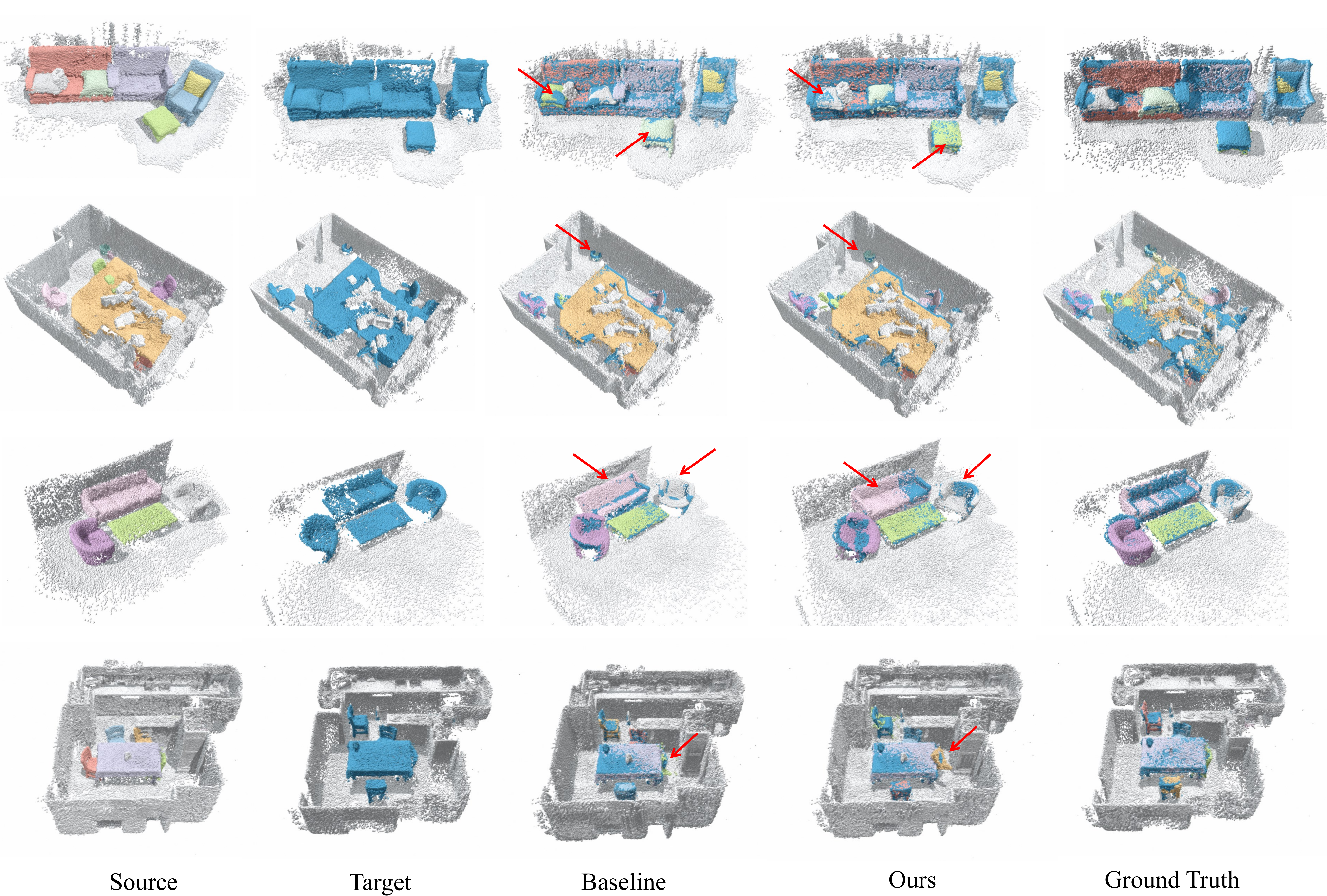}
    \caption{\textbf{Qualitative results of relocalization on \emph{3RScan}~\cite{wald2019rio}.} \more{} generates more correct matches thus better localization of moved instances. \boldsymbol{\textcolor{red}{$\searrow$}} highlights the differences between baseline and ours. In the first row, the baseline mismatched pillows with ottoman, leading to wrong registration. In the second row, our method outperforms the baseline on the trash can. In the third row, the baseline wrongly flipped the sofa colored in pink. In the last row, we present one scene with multiple identical chairs, where both methods fail in relocalizing all of them. Ours relocalizes only one and the baseline none of them.}
\label{fig:supp_vis_relocalization}
\end{figure*}

\begin{figure*}[t]
    \centering
    \includegraphics[width=\linewidth]{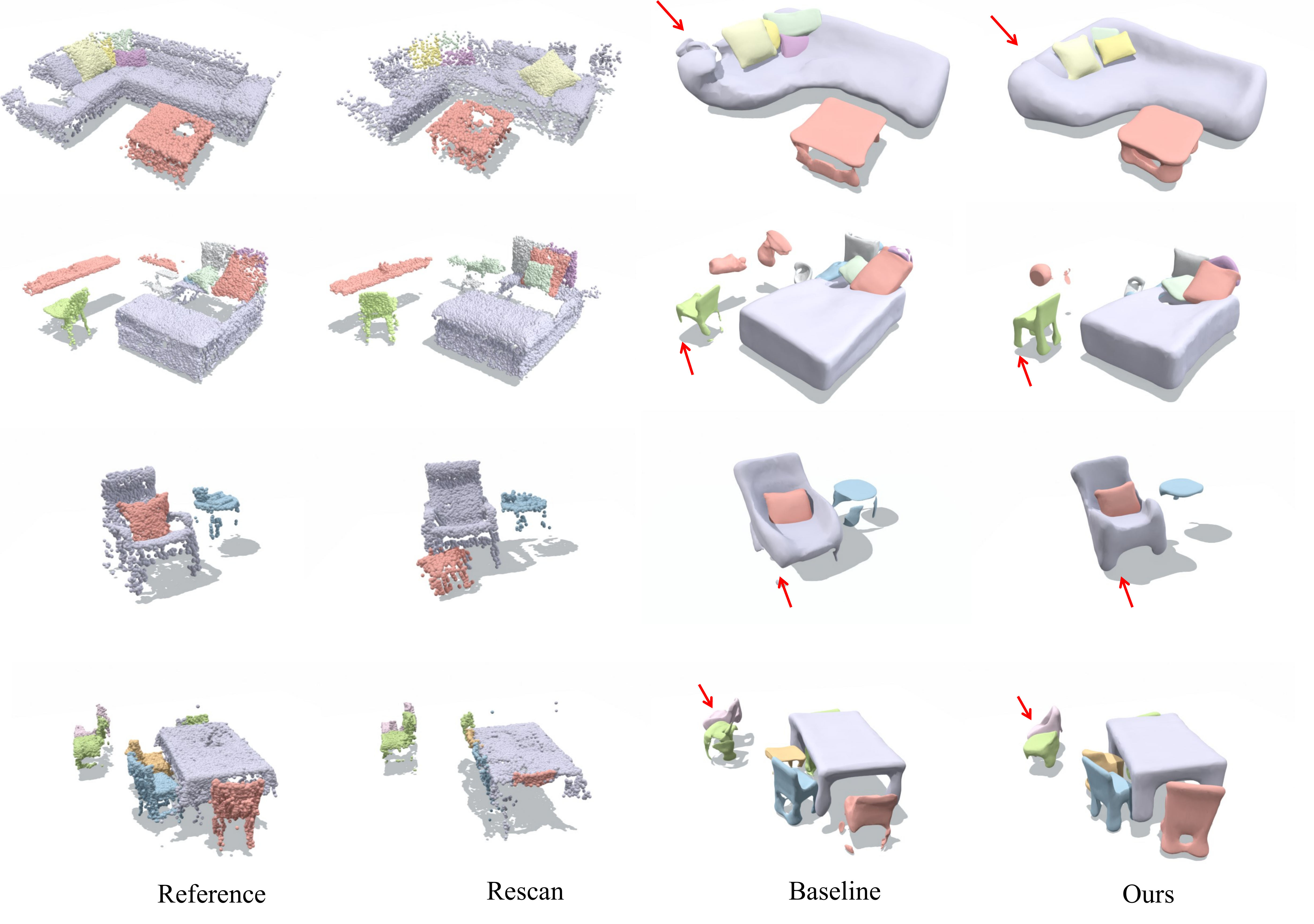}
    \caption{\textbf{Qualitative results of end-to-end performance on \emph{3RScan}~\cite{wald2019rio}.} Rescans are accumulated to the reference scan. The reconstruction of the scene is based on the accumulation, including the errors and noises provided from performing the task end-to-end. Compared to the baseline, our method is able to reconstruct cleaner and more complete surfaces by accumulating partial observations.}
\label{fig:supp_vis_recon}
\end{figure*}

We provide the following qualitative results: 
\textit{(1)} point cloud accumulation on \emph{FlyingShape} with GT input (\cf \cref{fig:supp_accumulation}); 
\textit{(2)} point cloud registration on \emph{FlyingShape} at instance level with GT input (\cf \cref{fig:instance_reg}); 
\textit{(3)} end-to-end multi-object relocalization on \emph{3RScan} (\cf \cref{fig:supp_vis_relocalization}); and
\textit{(4)} end-to-end multi-object relocalization and reconstruction (\cref{fig:supp_vis_recon}). In these additional visualizations, we provide more comprehensive qualitative evaluations, showing success and failure cases of our method and the baseline. The baseline method is the same per task as the one used in the main paper.

\section{Discussion}
\label{supp:discussion}
In this paper, we introduce \more{}, a novel method to parse long-term evolving environments. Through our extensive experiments on two datasets, we demonstrate the superior performance and robustness of our method even on partial, noisy point clouds with pose variations. 

\begin{figure*}[t]
    \centering
    \includegraphics[width=\linewidth]{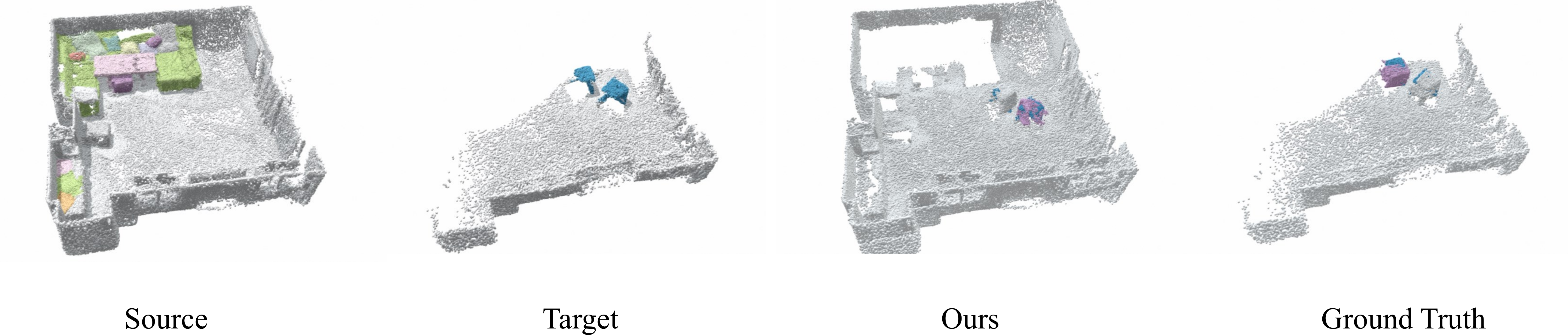}
    \caption{\textbf{A failure case of relocalization in \emph{3RScan}~\cite{wald2019rio}.} The failure is due to the large number of similar objects in the scene \ie, pillows, and the incomplete/low coverage of the scene in the rescan (target). As many instances in the source scan are no longer observed in the target scan, our method abandons unmatched ones as removed.}
\label{fig:failure_case}
\end{figure*}

\textbf{Limitations} \textit{(1)} Our method includes test-time optimizations and hence cannot run end-to-end in real-time. We consider it as a post-processing algorithm for temporal scans to understand the instance-level change (rigid motion and geometry) between them. The capture of indoor environment does not happen in real-time but instead has relatively long intervals. Therefore, real-time execution is not a necessity in this task. \textit{(2)} Our method faces challenges when dealing with multiple identical, similar, and/or symmetric shapes in the scene (\cf \cref{fig:failure_case}). This can be alleviated in future work by incorporating RGB-values and global context in the scene into our method. 

\end{document}